\documentclass[a4paper,fleqn]{cas-dc}

\usepackage[numbers]{natbib}
\usepackage{xcolor}
\usepackage{subcaption}
\usepackage{subcaption}
\usepackage{amsmath}
\usepackage{algorithm}
\usepackage{algpseudocode}
\usepackage{array}
\usepackage{pgfplots}
\usepackage{tabularray}
\usepackage{comment}
\usepackage{pifont}
\usepackage{float} 
\newcommand{\cmark}{\ding{51}}
\newcommand{\xmark}{\ding{55}}

\begin{document}
\let\WriteBookmarks\relax
\def\floatpagepagefraction{1}
\def\textpagefraction{.001}

\shorttitle{Axle Sensor Fusion for Online Continual Wheel Fault Detection in Wayside Railway Monitoring}

\shortauthors{Louren\c{c}o et~al.}

\title [mode = title]{Axle Sensor Fusion for Online Continual Wheel Fault Detection in Wayside Railway Monitoring}



%

\author[1]{Afonso Louren\c{c}o}[orcid=0000-0002-3465-3419]

\cormark[1]

\ead{fonso@isep.ipp.pt}

\author[1]{Francisca Osório}[orcid=0009-0001-2019-9251]

\cormark[1]

\ead{mfsgs@isep.ipp.pt}

\affiliation[1]{organization={GECAD, ISEP, Polytechnic of Porto},
    addressline={Rua Dr. António Bernardino de Almeida}, 
    city={Porto},
    postcode={4249-015}, 
    country={Portugal}}

\author[1]{Diogo Risca}[orcid=0009-0000-1495-0662]

\ead{difri@isep.ipp.pt}

\author[1]{Goreti Marreiros}[orcid=0000-0003-4417-8401]

\ead{mgt@isep.ipp.pt}

\cortext[cor1]{Equal contribution}


\begin{abstract}
Reliable and cost-effective maintenance is essential for railway safety, particularly at the wheel–rail interface, which is prone to wear and failure. Predictive maintenance frameworks increasingly leverage sensor-generated time-series data, yet traditional methods require manual feature engineering, and deep learning models often degrade in online settings with evolving operational patterns. This work presents a semantic-aware, label-efficient continual learning framework for railway fault diagnostics. Accelerometer signals are encoded via a Variational AutoEncoder into latent representations capturing the normal operational structure in a fully unsupervised manner. Importantly, semantic metadata, including axle counts, wheel indexes, and strain-based deformations, is extracted via AI-driven peak detection on fiber Bragg grating sensors (resistant to electromagnetic interference) and fused with the VAE embeddings, enhancing anomaly detection under unknown operational conditions. A lightweight gradient boosting supervised classifier stabilizes anomaly scoring with minimal labels, while a replay-based continual learning strategy enables adaptation to evolving domains without catastrophic forgetting. Experiments show the model detects minor imperfections due to flats and polygonization, while adapting to evolving operational conditions, such as changes in train type, speed, load, and track profiles, captured using a single accelerometer and strain gauge in wayside monitoring.

\end{abstract}


\maketitle

\section{Introduction}
\label{chap:Chapter1}

The railway sector relies on reliability and maintainability to ensure smooth and secure transport of passengers and freight \cite{davari2021predictive,lourencco2024time}. With trains designed to operate for decades and maintenance costs accounting for a significant portion of total expenditure, there is an urgent need to keep safety and service quality at an optimal level while minimizing operational costs \cite{ribeiro2016sequential,veloso2022metropt}. In particular, the wheel-track interface has received a lot of attention in the literature, as it incurs most of the cost of maintenance for both railway vehicles and infrastructure. These components are exposed to various environmental and operational conditions, which can induce corrosion, cracks, and other types of damage. In addition, surfaces of the wheel-track are subjected to high stick, sliding, and contact stresses in rolling contact.

\begin{figure}
\centering
\includegraphics[width=\linewidth]{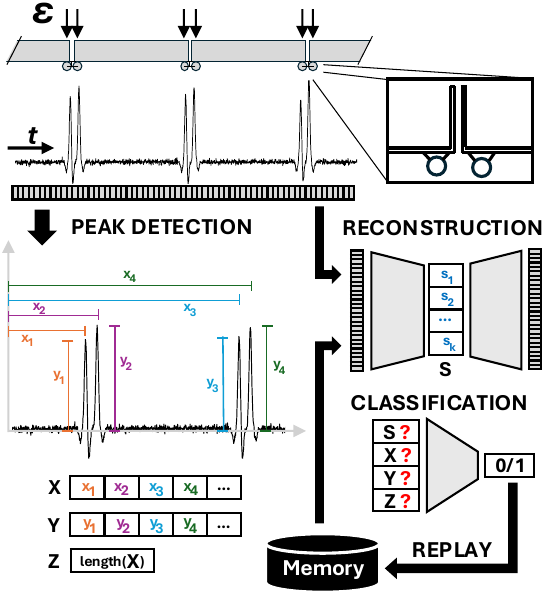}
\caption{(1) semantic extraction of temporal wheel indices (\textit{X}), strain-based deformation (\textit{Y}), and wheel count (\textit{Z}) via peak detection; (2) VAE-based reconstruction of signal embeddings (\textit{S}); and (3) fault detection through data fusion of \textit{S}, \textit{X}, \textit{Y}, and \textit{Z}, and continual loss-based experience replay.}
\label{fig:picgraphpaper}
\end{figure}

\textbf{Measurement systems.} To address this, predictive maintenance (PdM) strategies have emerged. Unlike traditional methods that involve scheduled checks and repairs, PdM leverages machine learning to predict potential equipment failures before they occur. This proactive approach not only enhances safety, but also optimizes maintenance resources by reducing unnecessary repairs and downtime, leading to significant cost reductions. Numerous measurement methods are used, such as laser-based systems \cite{jiang2019fast}, ultrasonic devices \cite{article1,gao2018use}, axle box acceleration \cite{MOLODOVA2016225,SALVADOR2016301}, eddy current \cite{gao2018use}, fiber Bragg grating \cite{Lai_2012}. Among these, time series have been identified as the most common format for sensors installed on trains and tracks \cite{lourencco2024time}.

\begin{figure*}
\centering
\begin{subfigure}{0.36\textwidth}
  \includegraphics[width=\linewidth]{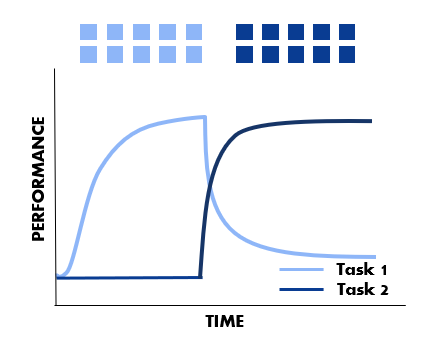}
  \caption{Sequential Training}
  \label{fig:catastforgetting1}
\end{subfigure}
\hspace{35pt}
\begin{subfigure}{0.36\textwidth}
  \includegraphics[width=\linewidth]{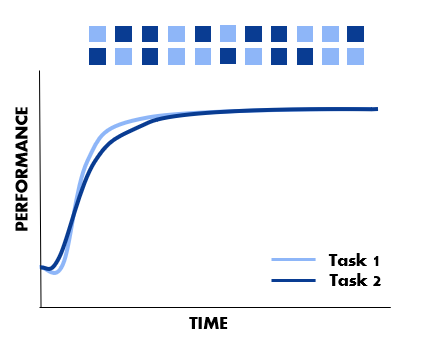}
  \caption{Interleaved Training}
  \label{fig:catastforgetting2}
\end{subfigure}

\caption{Sequential training causes rapid forgetting of the first task, while interleaved training allows the model to retain both, indicating that forgetting arises from the training strategy rather than model capacity.}
\label{fig:catast}
\end{figure*}

\textbf{Signal processing.} While these diverse measurement systems enable continuous monitoring of railway assets, the resulting raw signals are often high-dimensional, noise-contaminated, and heavily affected by operational variability, including traffic density, axle loads, and environmental conditions. Transforming such time series data into informative representations suitable for predictive maintenance therefore constitutes a critical intermediate step. To address this challenge, early machine learning (ML)–driven signal processing approaches largely adopted a task-specific paradigm, coupling carefully engineered signal processing techniques with ML models to isolate fault-related features. Representative methods include spectral kurtosis analysis \cite{s18051603,Mosleh2021}, Short-Time Fourier Transform (STFT) \cite{inproceedings12345}, Principal Component Analysis (PCA) \cite{LASISI2018230,SHARMA201834,LOURENCO2023107433}, envelope spectrum analysis \cite{s23042138,article2}, auto-regressive decomposition \cite{s23041910,doi:10.1080/00423114.2021.1912365}, Hidden Markov Models \cite{7094328, doi:10.1177/0954409713503460, doi:10.1177/1687814018806380}, wavelet transform \cite{6776541,s23041910}, and bivariate Gamma processes \cite{doi:10.1080/15732479.2011.563090}.

\textbf{Learned features.} However, as a substantial volume of data has been accumulated regarding the health status of machinery, diagnostic models should directly learn useful representations from raw data during deployment, regardless of the quality of training data. In this regard, the high degree of automation in data processing enabled by deep learning (DL) architectures, along with advances in model and dataset scalability, has marked a major leap forward, allowing models to handle extensive data features without requiring domain expertise. Additionally, it became evident that certain unstructured data types, such as time-series signals, could be effectively modeled by fine-tuning general-purpose DL models on specific datasets \cite{lourencco2025context,lourencco2025bridging}. As a result, research has increasingly shifted from manual feature engineering and shallow ML algorithms to multi-scale hierarchical latent representations of DL architectures for bearing fault diagnosis, such as stacked sparse autoencoders 
\cite{vamos}, convolutional neural networks \cite{faghih2016deep}, Markov Chain Monte Carlo-based Bayesian learning \cite{article3}, \cite{article4}, deep belief network \cite{yin2016fault}, and recurrent neural networks \cite{xu2020rail}. 

\textbf{Batch learning.} Although such DL methods can handle the diversity of fault types in a static dataset, these models are typically trained once, under the assumption that all possible fault types of the target equipment are fully covered during the training phase. However, in a real-world online setup, data is no longer shuffled but instead follows a natural sequence \cite{lourencco2025dfdt,neves2025online}, reflecting the actual progression of the conditions over time. This exposes the model to complex and dynamic conditions, with new faults, continuous equipment upgrades, and varying operational states emerging throughout their lifecycle. This presents a challenge to the traditional learning paradigm, which assumes that the availability of all training data is upfront and that the model will always exclusively encounter data originating from the same data distribution as encountered during training. To address this, one approach is to completely retrain the DL model, but this strategy requires significant storage capacity and is time-consuming, thus increasing operational and maintenance costs. Alternatively, retraining the model only with new data could be considered. However, since a model in this case only has access to the current dataset during each phase of the learning cycle, it is susceptible to overfitting the new data. Moreover, this method leads to catastrophic forgetting, where the model is prone to overfitting on the currently available data and suffers from performance deterioration on previous data. Figure \ref{fig:catast} illustrates this premise.

\textbf{Continual learning.} To address this, some continual learning (CL) techniques have been proposed for fault diagnosis, such as an adaptive deep learning framework designed to detect defects in high-speed railway that uses latent space and actively selects and annotates samples to update the training set \cite{9208746}, a statistical monitoring system for railway with ongoing data collection and analysis, where the model continuously updates its parameters based on new data \cite{HOFER20158557}, a few-shot learning model that extracts features from input images and encodes them into a latent space and employs a double-loop continual learning method, which simulates working memory and long-term memory processes \cite{10452127}, a boosting-inspired online learning approach with transfer learning that adapts to evolving operational conditions such as changes in speed, load, and track irregularities while mitigating catastrophic forgetting \cite{risca2025boosting}, or a model that utilizes a Transformer-based architecture with a graph encoder and temporal decoder to capture complex spatial and temporal dependencies and adapt to new speed patterns \cite{10400993}.

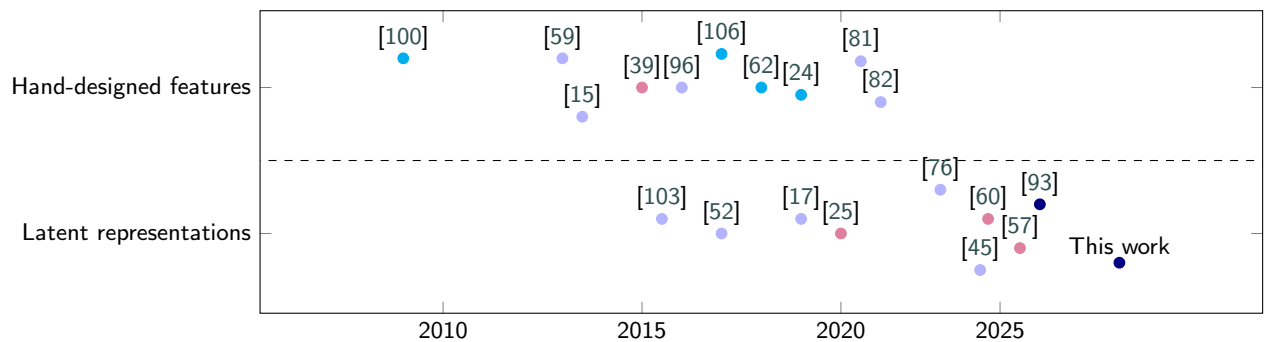
\begin{figure*}[h!]
    \centering
    \begin{tikzpicture}
    \begin{axis}[enlargelimits=0.2,
        xtick={2010, 2015, 2020, 2024},
        xticklabels={2010, 2015, 2020, 2025},
        ytick={0.5, 1.5},
        yticklabels={Latent representations, Hand-designed features},
        legend style={at={(1.05,1)}, anchor=north west},
        width=0.85\textwidth,
        height=0.32\textwidth]
        \draw[dashed] (axis cs:2004,1) -- (axis cs:2032,1);
        \addplot[
            scatter/classes={a={white!70!blue}, b={cyan}, c={black!50!blue}, d={purple!50!white}},
            scatter, mark=*, only marks, 
            scatter src=explicit symbolic,
            nodes near coords*={\Label},
            visualization depends on={value \thisrow{label} \as \Label} 
        ] table [meta=class] {
            x y class label
            2020 0.5 d {\cite{9208746}}
            2015 1.5 d {\cite{HOFER20158557}}
            2024.5 0.4 d {\cite{10452127}}
            2023.7 0.6 d {\cite{10400993}}
            2017 1.73 b {\cite{s17020318}}
            2019 1.45 b {\cite{s19163614}}
            2018 1.5 b {\cite{article7}}
            2009 1.7 b {\cite{article8}}
            2021 1.4 a {\cite{Mosleh2021}}
            2023.5 0.25 a {\cite{vamos}}
            2020.5  1.68 a {\cite{article2}}
            2015.5 0.6 a {\cite{7094328}}
            2017 0.5 a {\cite{article3}}
            2019 0.6 a {\cite{article4}}
            2022.5 0.8 a {\cite{s23041910}}
            2016 1.5 a {\cite{SALVADOR2016301}}
            2013 1.7 a {\cite{article}}
            2013.5 1.3 a {\cite{article1}}
            2025 0.7 c {\cite{risca2025boosting}}
            2027 0.3 c {This work}
        };
    \end{axis}
    \end{tikzpicture}
    \caption{Overview of related work: Offline/Batch processing (light blue), Online processing (blue), Offline Continual Learning (purple), Online Continual Learning (dark blue)}
    \label{fig:overview}
\end{figure*}

\textbf{Supervision dependence.} While effective, most of these methods highly depend on external descriptions and heavy supervision assumptions, failing when there are unknowns in testing, i.e. the open world setting \cite{chen2018lifelong}. These methods are narrowly focused on trade-offs surrounding catastrophic forgetting, e.g. inclusion of memory footprint, computational cost, cost of data storage, task sequence length and amount of training iterations, while disregarding the lack of supervision in a real-world online setting. In this regard, various online learning methods have been proposed for fault diagnosis, such as continual processing vibration signals through wavelet packet analysis \cite{s17020318}, parallelogram mechanisms for real-time monitoring \cite{s19163614}, a FBG-based online monitoring system \cite{article7}, an iterative search procedure for predicting flange contact points dynamically \cite{article8}, or a query-by-content approach used to obtain predictions, with Dynamic Time Warping (DTW) scores ranking the most anomalous wheel passages \cite{10.1145/3555776.3577860}. However, these don't rely on DL architectures, and thus don't face the challenge of catastrophic forgetting. Figure \ref{fig:overview} provides an overview.

\textbf{Domain semantics.} While it is important to develop methods that relax the supervision assumptions, in terms of both the definition of well defined boundaries and labeling of varying environmental and operational conditions, embedding the ability to leverage on such knowledge when its available can be extremely helpful. In this regard, railway monitoring systems present an excellent opportunity due to the wide practice of train detection signaling systems for proper railway traffic management \cite{Palmer_2010}. Instead of developing CL methods that rely on the availability of explicit domain boundaries, given in the form of external domain descriptors and hard-coded modular structures \cite{salehi2021unified,kim2023open}, semantic information, such as train load, speed, and wheel configuration, can be used as a proxy of the variations in operating conditions \cite{Raz, Ahmed_2020}.

\begin{figure*}[h!]
    \centering
    \begin{tikzpicture}
    \begin{axis}[enlargelimits=0.2,
        xtick={2007, 2009, 2011, 2013, 2015, 2017, 2019, 2021, 2023, 2025},
        xticklabels={2007, 2009, 2011, 2013, 2015, 2017, 2019, 2021, 2023, 2025},
        ytick={1, 3, 5},
        yticklabels={Optimizing Operations, Detection/Occupancy, Health Monitoring},
        ymin = 1, ymax =  5,
        legend style={at={(1.05,1)}, anchor=north west},
        width=0.85\textwidth,
        height=0.4\textwidth,
        ylabel near ticks]
        \draw[dashed] (axis cs:2007,2) -- (axis cs:2025,2);
        \draw[dashed] (axis cs:2007,4) -- (axis cs:2025,4);
        \draw[dashed] (axis cs:2007,6) -- (axis cs:2025,6);
        \addplot[
            scatter/classes={a={black!50!blue}, b={purple!50!white}, c={cyan}, d={purple!50!white}},
            scatter, mark=*, only marks, 
            scatter src=explicit symbolic,
            nodes near coords*={\Label},
            visualization depends on={value \thisrow{label} \as \Label} 
        ] table [meta=class] {
            x y class label
            2012.44 2.81 a {\cite{Gajda_2012}}
            2008.43 1.09 a {\cite{Midya_2008}}
            2019.43 4.2 a {\cite{Hidirov_2019}}
            2019.32 4.74 b {\cite{Liu_2019}}
            2017.37 4.2 b {\cite{Hu_2017}}
            2018.5 5.3 c {\cite{G_mez_2018}}
            2020.5 5.02 c {\cite{Luo_2020}}
            2015.43 4.59 c {\cite{Liang_2015}}
            2014.63 2.9 c {\cite{Palo_2014}}
            2019.55 5.3 b {\cite{Sun_2019}}
            2017.53  4.89 c {\cite{s17020318}}
            2016.54 4.95 c {\cite{SALVADOR2016301}}
            2018.17 4.12 c {\cite{Falamarzi_2019}}
            2012.4 5.13 b {\cite{Sadeghi_2012}}
            2013.47 4.92 c {\cite{article}}
            2014.12 4.26 c {\cite{Ye_2014}}
            2021.55 5.19 c {\cite{Meixedo_2021}}
            2023.51 5.19 c {\cite{LOURENCO2023107433}}
            2009.4 4.92 c {\cite{Ho_2009}}
            2014.56 5.17 c {\cite{Kinet_2014}}
            2011.56 4.65 c {\cite{Wei_2011}}
            2011.46 2.9 c {\cite{Filograno_2012a}}
            2010.48 2.97 c {\cite{Yan_2011}}
            2020.52 3.12 c {\cite{Zhang_2020}}
            2023.56 3.09 c {\cite{s23041910}}
            2015.43 0.91 c {\cite{Fumeo_2015}}
            2014.53 1.06 b {\cite{Li_2014}}
        };
    \end{axis}
    \end{tikzpicture}
    \caption{Related work: Traditional circuit detector (dark blue), Machine vision (purple), Vibration/optical fiber (light blue) }
    \label{fig:overview2}
\end{figure*}
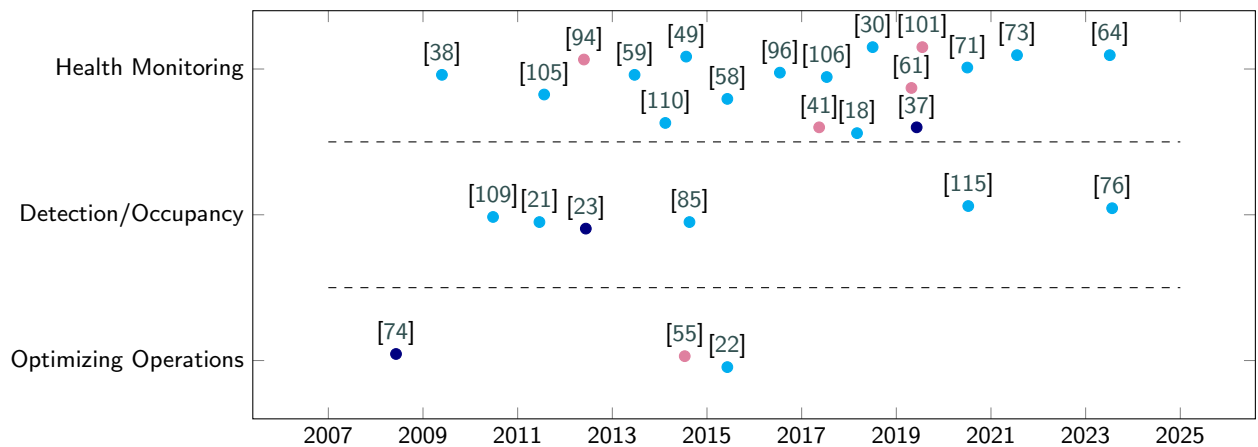

\textbf{Sensor-driven train characterization.} Historically, train characterization/detection technologies have been predominantly used in three main railway applications \cite{Hidirov_2019}. Firstly, to monitor the presence of trains in specific track sections, which is crucial for preventing collisions by maintaining safe distances between trains. Secondly, axle counters play a key role in infrastructure monitoring and anomaly detection, helping to detect irregularities such as unexpected stops or speed variations, which may indicate issues like track defects. Thirdly, when optimizing railway operations and traffic flow, axle counters provide real-time data on train positions, enabling efficient traffic management, reducing delays, and increasing network capacity. They also allow operators to make informed decisions regarding the regulation of train speeds, train movements and scheduling. Figure \ref{fig:overview2} provides an overview. Across these applications, train detection has been driven by both track circuits and axle counters \cite{Gajda_2012}. Track circuits operate as electrical systems with power sources and receivers placed along the track. These de-nergize when a train shunts the circuit, allowing for reliable detection of train presence. Axle counters, on the other hand, employ electromagnetic coils to detect wheels passing over the rail. By comparing the counts of axles entering and leaving a section of track, these systems provide accurate information on train presence and track occupancy.

\textbf{Eletromagnetic interference.} However, with the electrification of railways, electromagnetic interference (EMI) posed significant challenges. High-power electrical systems, such as traction drives, produce EMI that can disrupt the operation of track circuits and axle counters \cite{Midya_2008}. Moreover, while steps can been taken to separate the power sources for train detection systems and traction supplies, the difference in power levels might still lead to operational instability in highly electrified environments. To mitigate these challenges, ensuring electromagnetic compatibility (EMC) has become critical for the reliability of train detection systems. Standardized methods have been developed to quantify the radiated EMI from rail vehicles and measure the intensity of magnetic fields that affect track circuits and axle counters \cite{Bloomfield_2006}. These EMC standards and regulations can help ensure that systems can coexist without interference, however track circuits and axle counters are still prone to failure in hostile environments, such as those susceptible to flooding and high levels of EMI, driven by increasing train speeds and the higher power levels of modern propulsion systems \cite{Midya_2008}.

\textbf{Data-driven train characterization} To address these limitations, newer detection systems that are immune to EMI, such as laser-based sensors, have been introduced. These systems emit beams that are interrupted by passing train wheels, allowing for accurate axle counting without the risks posed by EMI. However, while effective, these systems are still being developed and implemented in limited railway networks. Alternatively, AI can also take a role in expanding the reliability of axle counting technologies for the optimizing railway operations, detecting infrastructure anomalies, and enhancing safety and security protocols within urban rail systems  \cite{Hidirov_2019}. Instead of specific sensors designed solely for train detection, AI allows to leverage general-purpose sensors to implicitly detect and count train axles, e.g. by machine vision paired with cameras to detect and analyze wheel patterns, such as flange height, flange thickness, and rim diameter, or vibration-based systems to monitor the forces exerted by train wheels on the track \cite{Y_ksel_2018, Chu_liang_Wei_2010, Buggy_2011, Allotta_2015}.

\textbf{Fiber Bragg grating.} For this purpose, Fiber Bragg Grating (FBG) sensors have proven successful across numerous applications, including measuring train speed, dynamic loads, and acceleration, as well as detecting derailment, rail cracks, and corrosion. FBG systems are immune to electromagnetic interference, lightweight, durable and responsive, providing high signal fidelity for variations in stress and strain \cite{Palo_2014, Falamarzi_2019, Falamarzi_2018, Falamarzi_2019, Sadeghi_2012, article}, temperature fluctuations \cite{Fumeo_2015, Rabatel_2011}, and pressure \cite{Li_2014}. These systems demonstrate strong potential for continuous monitoring and real-time assessment of railway infrastructure \cite{Kinet_2014}. It has even been shown that strain and stress signals from these sensors can be used directly for axle counting \cite{Li_2008}. Moreover, while FBG sensors are often preferred due to their ability to multiplex multiple signals along a single fiber, other optical fiber technologies have also been explored as alternatives to FBG, such as Brillouin-based sensors \cite{Yoon_2011, Minardo_2013}, intensity-based fiber optic sensors \cite{Chong_2014}, and even basic optical fuses \cite{Hopkins_2011}.

\textbf{Signal shape processing.} With this technology, the complexity of the measurement system has shifted from the  mechanics of sensor development to the peak detection process within the strain change signal as the weight of the train exerts significant force on the rail, in the vicinity of the measurement point. To capture this peak, various pre-processing techniques based on derivatives and correlation operations with a cut-off threshold have been proposed. For example, Weilai et al. introduced an FBG-based axle counter system, achieving 100\% accuracy in axle counting during over 1,000 consecutive tests in an experimental setting \cite{Y_ksel_2018}. Similarly, \cite{Chu_liang_Wei_2010} developed the X-Crossing and D-Crossing algorithms to determine the number of axles passing through a measurement station by using a cut-off threshold followed by a derivative operation to extract the relevant peaks in the signal data. These systems have been thoroughly validated through field testing, with some expanded for use in real-time wheel defect detection \cite{Wei_2011}. Other implementations, such as \cite{Buggy_2011}'s axle counting system, also rely on threshold and derivative-based operations, while \cite{Ho_2009} employed wavelet transforms to decompose signals into frequency bands, studying the spectral characteristics of the useful signals. \cite{Allotta_2015} developed a correlation-based algorithm for estimating various parameters under different vehicle and track configurations. Furthermore, these FBG-based methods have been extensively tested in high-speed railway systems for real-time strain measurement, axle counting, and train identification \cite{Filograno_2012a, Yan_2011}, and have even been adapted for wheel flat detection \cite{Filograno_2013a}. In light of these advancements, a self-checking system using FBG has also been proposed to detect sensor loosening. This system uses FBG to monitor the pre-tightening force of mounting bolts, detect the installation status of rail-mounted devices, and send an alarm if looseness is detected \cite{Zhang_2020}.

\textbf{This work.} Building on these advancements, this work presents a comprehensive study of the potential of semantic contextual information, via signal processing of vibration-based sensors, to enrich the latent representations of a continual learning system. While the proposed CL method does not require knowledge of task identity and boundaries at any point, eliminating the dependency on an oracle for domain information to adapt their learning process, it can still leverage on the semantic information to better handle the shifts in data distribution, as well as, better forward and backward transfer upon the repetitive reappearance of domains with similar environmental and operational conditions.

\begin{figure*}
\centering
\includegraphics[width=\linewidth]{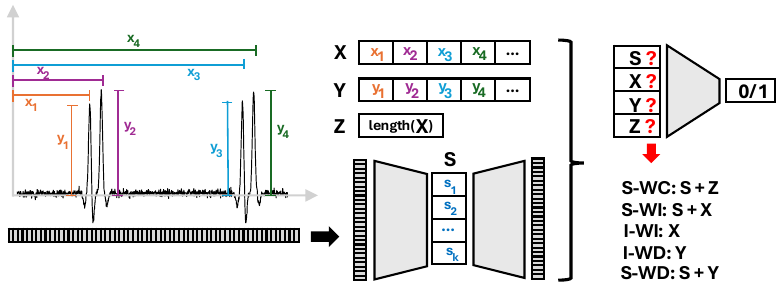}
\caption{Semantics extraction of temporal wheel indices (\textit{X}), strain-based deformation (\textit{Y}), and wheel count (\textit{Z}) via peak detection, which are subsequently fused with VAE-based reconstruction of signal embeddings (\textit{S}), under an ablation of five different strategies: \textit{S-WC} = \textit{S} + \textit{Z}, \textit{S-WI} = \textit{S} + \textit{X}, \textit{I-WI} = \textit{X}, \textit{I-WD} = \textit{Y}, and \textit{S-WD} = \textit{S} + \textit{Y}. Further details in Table \ref{tab:ad_summary}.}
\label{fig:picgraphpaper2}
\end{figure*}

\section{Methodology}
\label{sec:methodology}

The proposed fault diagnosis methodology encompasses three main components: (1) semantic extraction; (2) reconstruction of signal embeddings; and (3) fault detection through data fusion, and continual loss-based experience replay. Figure \ref{fig:picgraphpaper} provides an overview. 

\begin{figure}
\centering
    \begin{subfigure}{0.2\textwidth}
    \includegraphics[width=\linewidth]{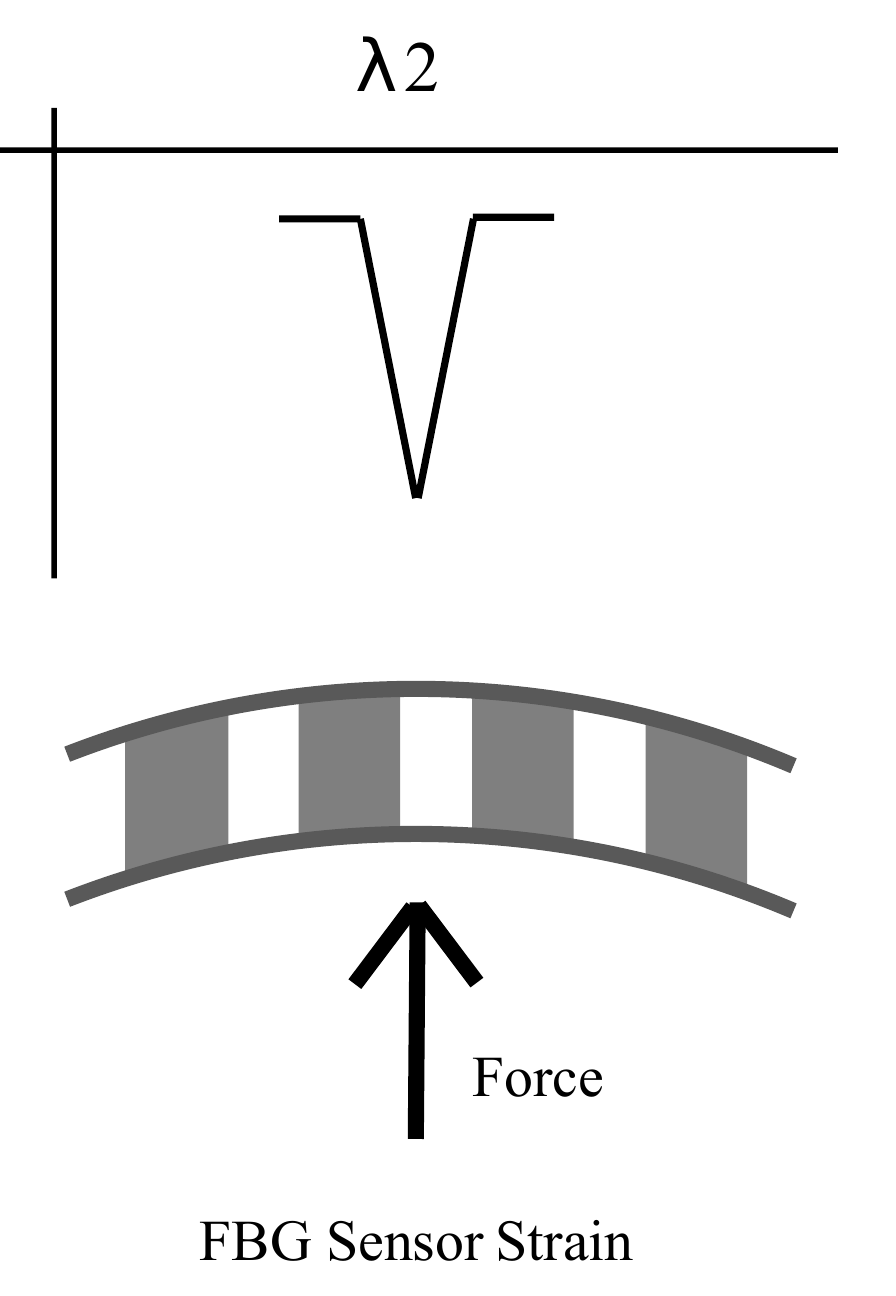} 
      \caption{Positive}
      \label{fig:emptyscheme}
    \end{subfigure}
    \begin{subfigure}{0.2\textwidth}
    \includegraphics[width=\linewidth]{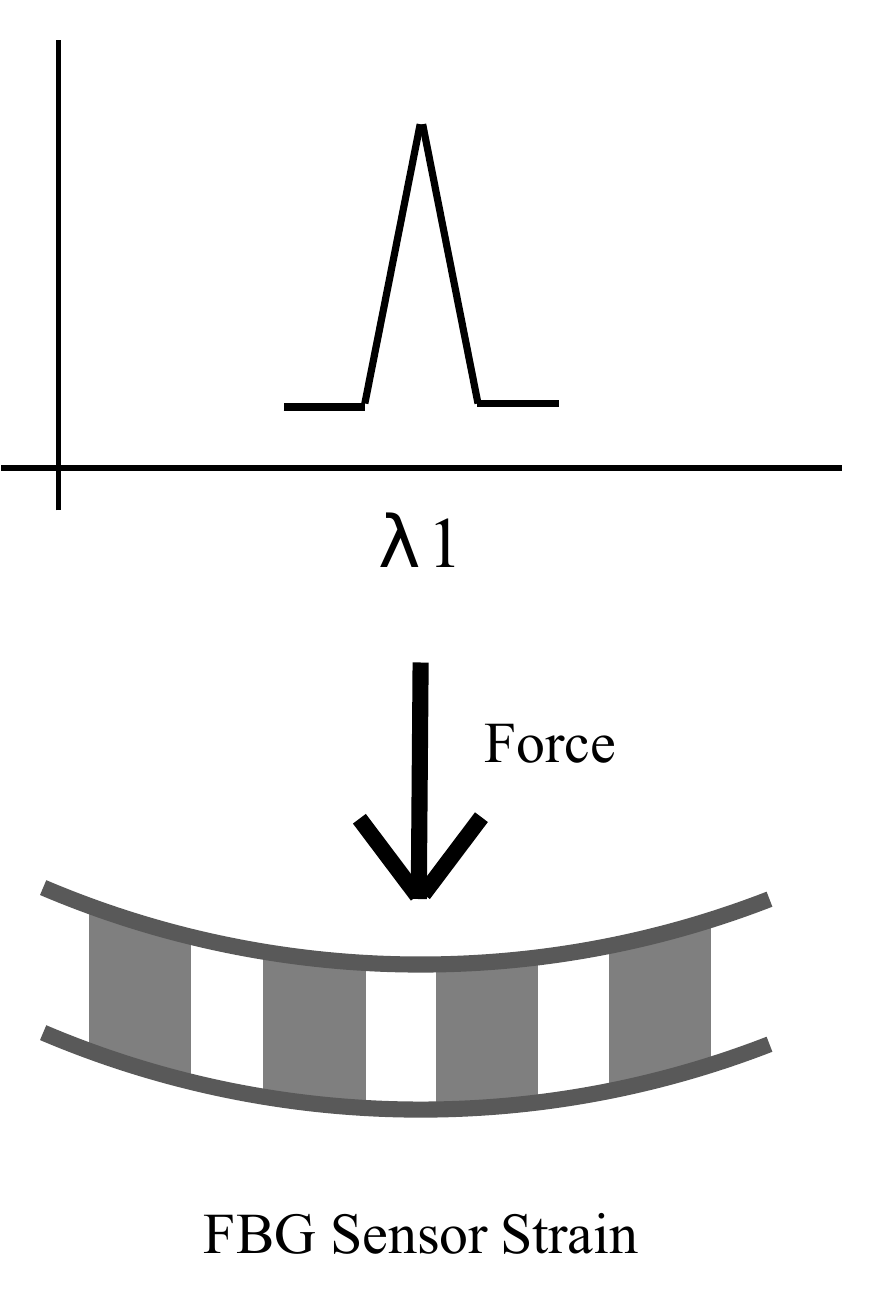} 
      \caption{Negative}
      \label{fig:halfscheme}
    \end{subfigure}
    \caption{Reflective wavelengths}
    \label{fig:Reflectivewavelengths}
\end{figure}

\textbf{FBG-driven train characterization.} Leveraging on vibration-based sensors, such as FBG and strain gauges, allows to detect the strain changes in the rail as trains pass over specific measurement points. When a train traverses these points, the rail experiences momentary deformation, with the rail head undergoing compression and the rail foot experiencing tension, as illustrated in Figure \ref{fig:Reflectivewavelengths}. According to the literature, the maximum absolute value of the signal's extension offers a more reliable and accurate criterion for detecting axles, rather than the signal length. By counting the number of peaks generated from each sensor, FBG sensors can effectively perform the axle counting function. With a single FBG sensor, the speed of the train can be accurately determined, assuming the train configuration is known. By positioning two sensors along the track, both the speed and direction of the train can be measured, assuming the distance between them is known. With four sensors strategically placed on the rail web, the system can also provide information about the axle load \cite{Kouroussis_2016}.

\textbf{Axle count semantics.} Different experiments were performed, each one focusing on different levels of semantic information extracted from the axle count. S-WC relies solely on the number wheels counted, while S-WI incorporates the precise temporal indexes of wheels in addition to their count. For ablation purposes, two experiments rely exclusively on the extracted semantics, ignoring corresponding VAE learned representation. I-WI uses only the timing and counting data from axle detections, while I-WD uses deformation values captured by strain sensors during each wheel passage. Finally, S-WD integrates strain signal deformations with semantic information about axle conditions. An overview of the experimental framework for semantics extraction is presented in Figure~\ref{fig:picgraphpaper2}, with further details in Table \ref{tab:ad_summary}. Beyond accelerometer signals, the model could also integrate semantic information, such as train load and speed, to enhance predictive performance \cite{Raz, Ahmed_2020}. To study their impact, it is important to note that each of these experiments was conducted under two distinct scenarios. In Scenarios S-WC, S-WI, I-WI, I-WD, and S-WD, the models were trained using only the described setup. In contrast, Scenarios S-WC*, S-WI*, I-WI*, I-WD*, and S-WD* extended the model's semantic understanding by incorporating additional context, specifically, information on train load and speed.

\begin{table*}[h!]
\centering
\caption{Anomaly detection (AD) experiments: semantic context, embedding use, and key focus}
\begin{tabular}{lllp{11.5cm}}
\toprule
\textbf{Code} & \textbf{Semantic} & \textbf{Signal} & \textbf{Description} \\
\midrule
\textbf{S-WC} & Wheel count & \cmark & Detects anomalies in wheel count using strain signals and embeddings. Flags missing wheels or signal degradation. \\
\hline
\textbf{S-WI} & Wheel indexes & \cmark & Adds temporal wheel indexes to detect spacing irregularities, misalignments, improper bogie assembly, or mechanical faults. \\
\hline
\textbf{I-WI} & Wheel indexes & \xmark & Uses only axle timing and counting to detect abnormal patterns, such as missing axles, or misalignments. \\
\hline
\textbf{I-WD} & Axle deformations & \xmark & Analyzes strain-based deformations, implicitly identifying uneven loading or undercarriage faults. \\
\hline
\textbf{S-WD} & Axle deformations & \cmark & Combines axle deformation signals with semantic axle context to capture subtle, early-stage anomalies. \\
\bottomrule
\end{tabular}
\label{tab:ad_summary}
\end{table*}

\textbf{Signal shape processing.} Naturally, these measurable reflective responses of strain and stress changes can be affected by the deterioration of wheel/rail contact conditions. Thus, to evaluate the extent to which FBG sensors can provide a reliable method for axle counting, train speed, direction determination, axle load measurement, four different peak detection techniques were compared: \texttt{detect\_peaks\_tony}, \texttt{peakdetect}, \texttt{detect\_peaks}, and \texttt{find\_peaks}. These have the same working principle, by locating the local maxima in a one-dimensional signal array, which correspond to the positions where each wheel exerts maximum strain on the sensor, i.e. a peak is detected at position $n_i$ if $x[n_i] > x[n_i - 1] \quad \text{and} \quad x[n_i] > x[n_i + 1]$ for a discrete signal $x[n]$. Moreover, these functions can be configured to detect peaks that are higher than their immediate neighbors by a specified prominence. The prominence $P$ of a peak at position $n_i$ measures how much the peak stands out due to its intrinsic height and its location relative to other peaks, i.e., $P(n_i) = x[n_i] - \max\left( \min(x[L], x[R]) \right)$, where $L$ and $R$ are the indices of the left and right bases of the peak, respectively. While similar in their purpose, these methods employ distinct computational strategies. The \texttt{detect\_peaks\_tony} algorithm normalizes the signal using the root mean square (RMS), computes peak-to-average power ratios, i.e., $\left( \text{signal}[i] \right)^2 / \text{RMS}$, to enhance peak prominence, and applies logical conditions to identify points exceeding their neighbors and a specified threshold, returning the indices of these peaks. The \texttt{peakdetect} method identifies local maxima and minima by iteratively comparing signal values within a defined lookahead distance, applying a delta threshold to confirm significant extrema, and producing lists of minima and maxima indices. The \texttt{detect\_peaks} function analyzes the signal's first-order differences to identify transitions indicative of peaks, applying a minimum amplitude threshold for filtering and outputting the corresponding indices. The \texttt{find\_peaks} function, part of the \texttt{scipy.signal} library, utilizes a prominence metric to quantify how distinctly a peak stands out relative to its surroundings, filtering based on this criterion and returning indices of prominent peaks.

\begin{figure*}
    \centering
    \includegraphics[width=\textwidth]{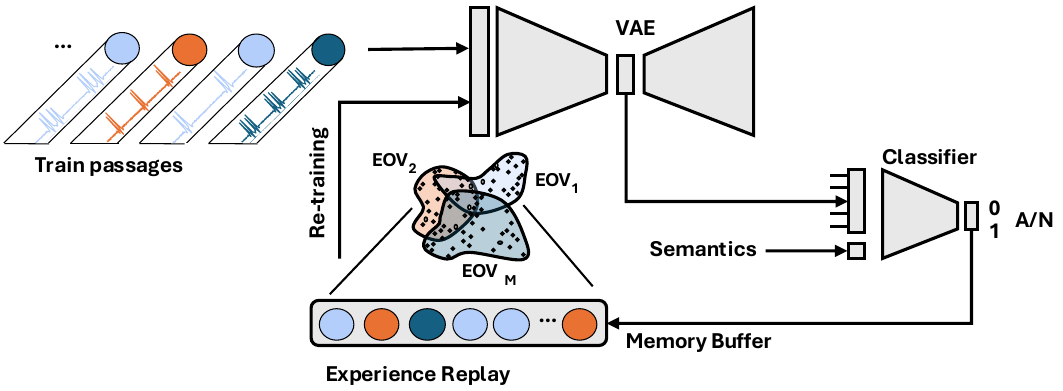}
    \caption{Fault detection through continual loss-based experience replay}
    \label{fig:diagramatot}
\end{figure*}

\textbf{Signal compression.} To encode accelerometer signals into compact latent representations that capture normal operational behavior in a fully unsupervised manner, a Variational Autoencoder (VAE) is employed \cite{Ma_2020, Lee_2019, Jiang_2017}. As a generative model, the VAE approximates the unknown data-generating distribution \( p(x) \) of an observed random vector \( x \) by learning a structured latent space that preserves essential signal characteristics while filtering noise, a property shown to be effective for subtle, context-dependent anomaly detection \cite{8682702}. Assuming i.i.d. samples, the marginal log-likelihood of the dataset, \(\ln p(x_1, \dots, x_N)\), can be expressed as a sum over individual samples $\sum_{i=1}^N \ln p(x_i)$, the VAE introduces a latent variable \( z \in \mathbb{R}^d \) that serves as a latent representation of \( x \). The generative process is described by:  
\[
p_\theta(x) = \int_{\mathbb{R}^d} p_\theta(x \mid z) p_z(z) \, dz, \tag{1}
\]  
where \( p_z(z) \) is the prior distribution over the latent space \( \mathbb{R}^d \), typically chosen as a multivariate standard Gaussian \( \mathcal{N}(0, I_d) \). The conditional distribution \( p_\theta(x \mid z) \), referred to as the decoder, is parameterized by a neural network. Since the posterior \( p_\theta(z \mid x) \) defined via Bayes' rule as $p_\theta(z \mid x) = \frac{p_\theta(x \mid z)p_z(z)}{p_\theta(x)}$ is generally intractable, variational inference is needed \cite{graves2011practical}. To approximate the posterior, the VAE introduces an encoder network that models an approximate posterior \( q_\phi(z \mid x) \), also parameterized by neural networks and typically modeled as \( q_\phi(z \mid x) = \mathcal{N}\left(z; \mu(x), \Sigma(x)\right) \), where \( \mu(x) \) and \( \Sigma(x) \) are outputs of the encoder network. Using importance sampling with $q_\phi(z|x)$, an unbiased estimate $\hat{p}_\theta$ of the marginal likelihood $p_\theta(x)$ is given by:  
\[
\hat{p}_\theta(x) = \frac{p_\theta(x \mid z)p_z(z)}{q_\phi(z \mid x)}, \quad \text{with} \quad \mathbb{E}_{z \sim q_\phi} \left[\hat{p}_\theta(x)\right] = p_\theta(x).
\]  

with \( \phi \) and \( \theta \) denoting the parameter sets modeled respectively by the decoder and encoder neural networks. Applying Jensen's inequality to the marginal likelihood results in the Evidence Lower Bound (ELBO):

\[
\ln p_\theta(x) \geq \mathbb{E}_{z \sim q_\phi(z \mid x)} \left[ \ln p_\theta(x \mid z) \right] - \text{KL} \left[q_\phi(z \mid x) \| p_z(z)\right]. \tag{2}
\]  

The ELBO serves as the loss function for training the VAE, consisting of two terms: the reconstruction loss and the regularization loss. The reconstruction loss maximizes the expectation of the log-likelihood \( \ln p_\theta(x \mid z) \) with respect to the approximate posterior \( q_\phi(z \mid x) \), ensuring that the decoder accurately reconstructs the input \( x \). The regularization loss minimizes the Kullback-Leibler (KL) divergence between the approximate posterior \( q_\phi(z \mid x) \) and the prior \( p_z(z) \), aligning the latent variable distribution with the prior, which is typically chosen as a simple isotropic Gaussian. Since the KL divergence is non-negative, optimizing the ELBO drives \( q_\phi(z \mid x) \) closer to \( p_z(z) \) while simultaneously improving reconstruction accuracy.

\textbf{Continual learning.} Building on recent findings that semi-supervised anomaly detection methods \cite{pang2019deep,goyal2020drocc} can outperform fully unsupervised approaches even when only 1\% of anomalous instances are available \cite{han2022adbench,pauperio2025explainable}, the proposed framework avoids relying directly on reconstruction errors, which are often overly sensitive to minor perturbations in evolving operational conditions. Instead, a lightweight supervised classifier, i.e. XGBoost, is trained on the learned embeddings to provide a more stable and robust decision layer. Moreover, unlike traditional anomaly detection which is performed on a static dataset, continual anomaly detection with evolving domains needs to adapt across dynamically changing domains without forgetting \cite{lourencco2025device}. The objective is to minimize the total risk across all domains, i.e., $ L_t(\theta) + L_{1:t-1}(\theta) = \mathbb{E}_{(x,y) \sim D_t}[\ell(y, h(x; \theta))] + \sum_{i=1}^{t-1} \mathbb{E}_{(x,y) \sim D_{i}}[\ell(y, h(x; \theta))]$, where $L_t$ calculates model $h_\theta$'s expected prediction error $\ell$ over the current domain's data distribution $D_t$. $L_{1:t-1}$ is the total error evaluated on the past $t - 1$ domains' data distributions, i.e., $\{D_i\}_{i=1}^{t-1}$. The main challenge of continual learning comes from the practical memory constraint that no (or only very limited) access to the past domains' data is allowed. Under such a constraint, it is difficult, if not impossible, to accurately estimate and optimize the past error $L_{1:t-1}$. Therefore, the main focus of recent domain incremental learning methods has been to develop effective surrogate learning objectives for $L_{1:t-1}$. Among these, replay-based methods, which reinforce the importance of old concepts interleaved in consecutive training, i.e. assuming stored exemplars are prototypes of the past data distributions, have been shown to perform extremely well \cite{prabhu2020gdumb}. By repeatedly sampling from this buffer, the model breaks the temporal correlation that characterizes streaming and sequential data and helps in stabilizing its understanding of normal behavior and improving the convergence of the model. Thus, correcting for the short-term bias towards recent data in the VAE objective function to the extent that the past is a good proxy for the future.

\textbf{Experience replay.} One typical example is Experience Replay (ER) \cite{riemer2019learning}, which stores a set of exemplars $M$ and uses a replay loss $L_{\text{replay}}$ as the surrogate of $L_{1:t-1}$. In addition, a fixed, predetermined coefficient $\beta$ is used to balance current domain learning and past sample replay, represented as: $L_t(\theta) + \beta \cdot L_{\text{replay}}(\theta) = L_t(\theta) + \beta \cdot \mathbb{E}_{(x',y') \sim M}[\ell'(y', h(x'; \theta))]$. To minimize this gap between the surrogate loss ($\beta \cdot L_{\text{replay}}$) and the true objective ($L_{1:t-1}$), there are 3 key ER components that can be tuned: rehearsal representation, label strategy, and rehearsal policy. Figure~\ref{fig:diagramatot} provides an overview of the complete system. 

\begin{figure*}[t]
    \centering
    \includegraphics[width=0.95\textwidth]{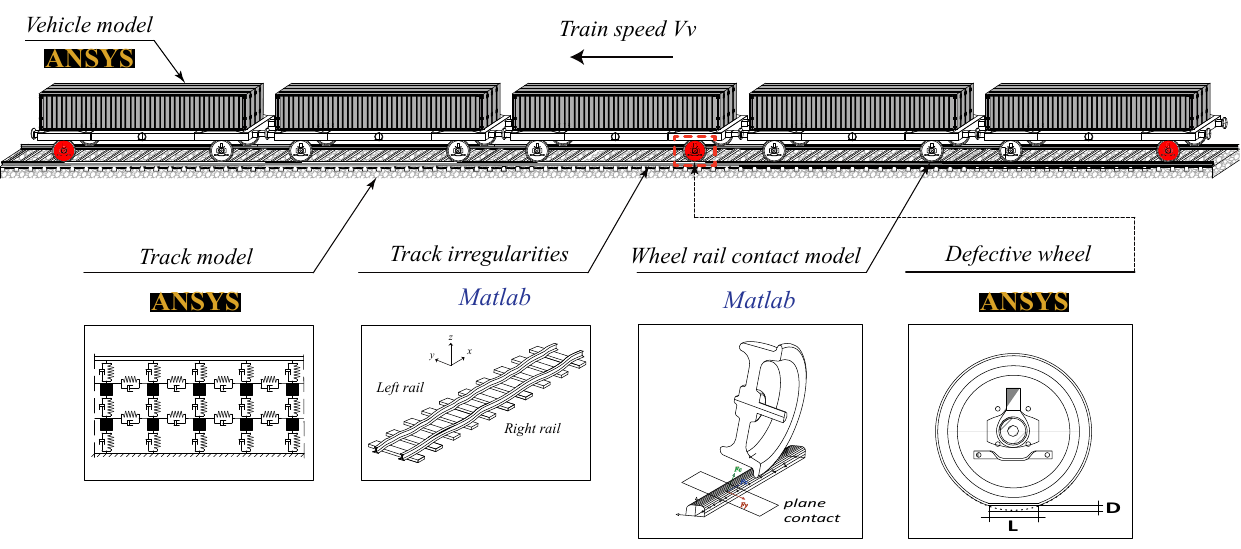}
    \caption{Train-track dynamic interaction}
    \label{fig:train_track_dynamic}
\end{figure*}

\begin{itemize}
    \item \textbf{rehearsal representation:} one can store raw input data and latent features from hidden layers on a memory buffer \cite{hayes2021remind}. For instance, one can replay compressed representations of activations (network embedding) of the intermediate layers in addition to the input-output pairs \cite{caccia2022new}. Alternatively, leveraging on the ability to generate new data from the prior distribution of latent variables, one can also produce pasts instances synthetically. This pseudo-rehearsal based method allows to save old concepts in modeling parameters instead of a memory buffer for later rehearsal \cite{shin2017continual}. In this work, we used latent embeddings as a more rich and informative sample of the signal's underlying structure and characteristics, avoiding the cumbersome long raw signals.
    \item \textbf{label strategy:} the memory buffer can hold the true label, which is not always be feasible in real-time applications, or a predicted label by the model. For instance, one can use additional logits distillation, referred to as dark experience replay \cite{buzzega2020dark}. In this work, we adopted the fundamental memory-based domain incremental learning framework where the mini-batch of the memory is regularly replayed with the current domain data \cite{riemer2019learning} as well as soft labels \cite{buzzega2020dark}.
    \item \textbf{rehearsal policy:} one can identify which stored samples should be selected for rehearsal using a variety of sampling policies e.g., uniform balanced, min rehearsal, max loss, min margin, min logit-distance, and min confidence \cite{hayes2021replay,chaudhry2018efficient,harun2023online,prabhu2020gdumb}. Alternatively, MIR make virtual updates using the incoming data to find the maximally interfered old samples for a loss-based policy \cite{aljundi2019online}. In this work, both reservoir sampling and a loss-based approach were used. Reservoir sampling ensures that the buffer maintains a representative subset of the entire data distribution, where a new sample overwrites a randomly selected sample from the buffer once the buffer is full, where the probability that item \( m \) remains after processing all \( n \) items is given by the product $\left(1 - k / (m+1)\right) \times \left(1 - k / (m+2)\right) \times \cdots \times \left(1 - k / n\right))$, which simplifies to ${k}/{n}$. Loss-based sampling strategy prioritizes the retention of informative and challenging data points \( D_{\text{retain}} \)  whose loss \( \ell(f(x; \theta), y) \) exceeds a threshold \( \tau \).
\end{itemize}

\section{Wheel-rail simulation}
\label{sec:case}

\begin{figure}[h]
    \centering
    \includegraphics[width=0.95\linewidth]{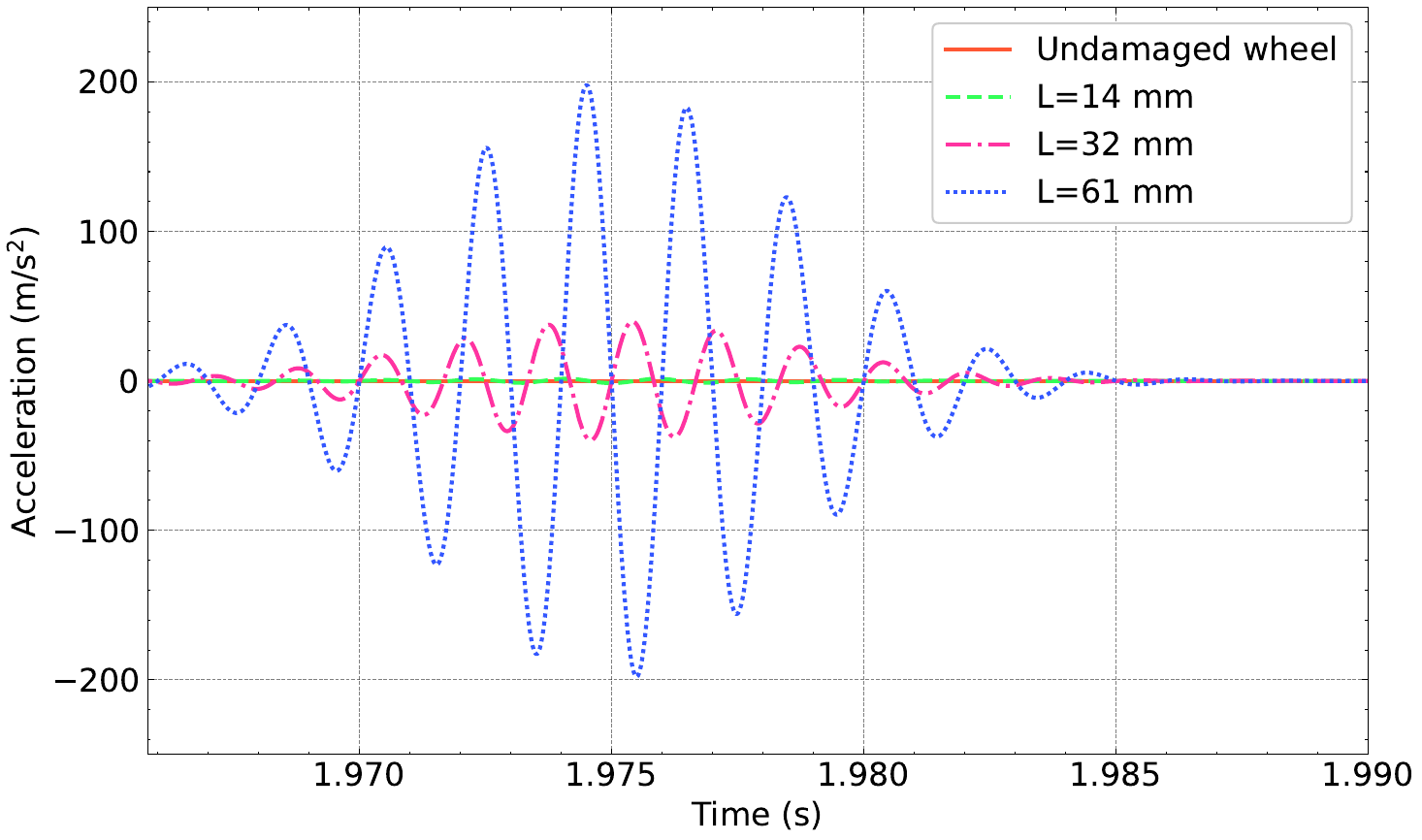}
    \caption{Acceleration for different flat lengths}
    \label{fig:acc_flats}
\end{figure}

This section describes the modeling and simulation process of wheel-rail interaction across various operational scenarios, while inducing minor imperfections, such as polygonization and flats (Figure \ref{fig:acc_flats}), captured with a single accelerometer and strain gauge positioned along the track.

\textbf{Train-track dynamic interaction.} To investigate train-track dynamic interaction, validated numerical simulations were employed. These simulations used VSI software, previously described and validated by the authors in a previous publication \cite{MONTENEGRO2015200}. The VSI model established a coupling between the train and track through a sophisticated 3D wheel-rail contact model (Figure \ref{fig:train_track_dynamic}). This model incorporates Hertzian contact theory to compute normal contact forces, while the USETAB routine assesses tangential forces arising from rolling friction creep phenomena \cite{Hertz+1882+156+171,kalker1996book}. Notably, the VSI software operates within the MATLAB® environment, enabling the import of structural matrices from both the train and track structure. These structures were previously modeled independently using a finite element (FE) package. The track structure within the simulations was meticulously represented using beam elements for rails and sleepers. Additionally, spring-dashpot elements were employed to capture the behavior of flexible layers, such as ballast and fasteners/pads. Finally, mass point elements were included to account for the mass of the ballast. The train model, constructed within ANSYS®, utilized a multibody formulation. This formulation incorporates spring-dashpot elements to simulate the flexibility of primary and secondary suspensions, rigid beams to represent the vehicle's rigid body movements, and mass point elements positioned at the center of gravity of each body (carbody, bogies, and wheelsets) to capture their mass and inertial effects.

\begin{table*}[ht]
    \centering
    \caption{EOVs for domain division}
    \label{tab:comparision_baseline_damage}
    \begin{tabular}{l c c c} 
    \toprule
         & \textbf{Baseline} & \textbf{Flat} & \textbf{Polygonized} \\ 
    \midrule
        \textbf{Train load} & 6 (full, half, empty, and 3 diff. unbalanced) & 1 (full) & 1 (full) \\
        \textbf{Irregularity profiles} & 96 & 24 & 24 \\
        \textbf{Train speeds (km/h)} & 40-220 & 60-200 & 60-200 \\
        \textbf{Defect locations} & - & 3rd wagon, 1st/3rd left & 1st wagon, 1st right \\
        \textbf{Amplitude (mm)} & - & \makecell[c]{10-20, 25-50, 40-100} & \makecell[c]{.25-.35, 0.55-.75} \\
        \textbf{Depth (mm)} & - & \makecell[c]{.02-.06, .09-.16, .23-.36} & \makecell[c]{6-8, 12-14, 17-20, 29-30} \\
    \bottomrule
    \end{tabular}
\end{table*}

\begin{figure*}
    \centering
    \begin{subfigure}{0.46\textwidth}
        \includegraphics[width=\linewidth]{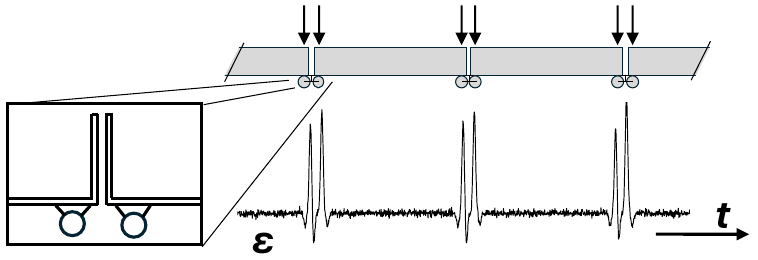}
        \caption{Laagrss vehicle}
    \end{subfigure}\hfill
    \begin{subfigure}{0.5\textwidth}
        \includegraphics[width=\linewidth]{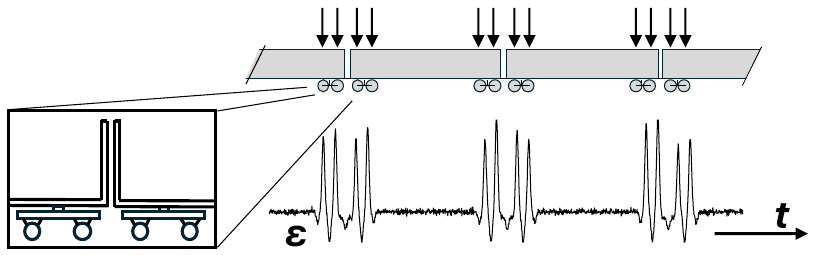}
        \caption{Alfa vehicle}
    \end{subfigure}
    \caption{Axle configurations across train types}
    \label{fig:traintypes}
\end{figure*}

\textbf{Numerical wayside modeling.} An accelerometer and strain gauge are positioned along the rail for the out-of-roundness detection system, designed to capture wheel defects. To validate the methodology, comprehensive 3D simulations based on train-track interaction were performed. Virtual simulations of both baseline and damaged wheels were conducted to validate the automatic out-of-roundness diagnosis. Once the system was validated, it demonstrated the potential to identify different types of trains and various wheel defects. It is also suitable for use with real-world data, where wheels often exhibit imperfections, such as flats or polygonal damage, that affect wheel-rail contact forces and cause vibrations in both the train and track components. Table \ref{tab:comparision_baseline_damage} provides a detailed comparison between the baseline and damage domains, resulting from combinatorial exploration of the aforementioned EOVs and defect types. Two types were considered: wheel flats appear either on the 1st our 3rd left wheel of the 3rd wagon, while polygonization appears on the 1st right wheel on the 1st wagon. 

\textbf{Flats.} For wheel flats, two flat length intervals (\(L_w\)) were considered, designated as L1 and L2. Figure~\ref{fig:acc_flats} illustrates the effect of different severities of wheels flat. The uniform distributions U (25, 50) mm and U (50,100) mm define the lower and upper limits of the flat length of the wheel for each interval L1 and L2, respectively. The wheel flat depth (\(D_w\)) is calculated on the basis of equation \cite{article9}: \(D_w = \frac{L_w^2}{16R_w}\), in which \(R_w\) is the radius of the wheel. The vertical profile deviation of the wheel flat is defined as:
\begin{equation}
\scalebox{0.8}{$
\begin{aligned}
    -\frac{D_w}{2} \left( 1 - \cos \frac{2\pi x_w}{L_w} \right) \cdot 
    H\Big(x_w - \big(2\pi R_w - L_w\big)\Big),  0 \leq x \leq 2\pi R
\end{aligned}
$}
\end{equation}

where \(H\) represents the Heaviside function and \(x_w\) is the coordinate aligned with the longitudinal direction of the track.

\begin{figure*}
    \centering
    \begin{subfigure}{0.33\textwidth}
        \includegraphics[width=\linewidth]{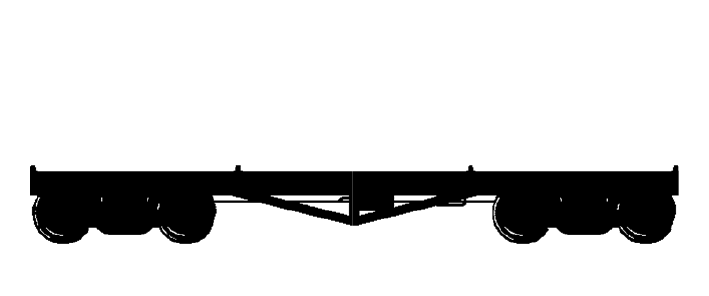}
        \caption{Empty}
    \end{subfigure}\hfill
    \begin{subfigure}{0.33\textwidth}
        \includegraphics[width=\linewidth]{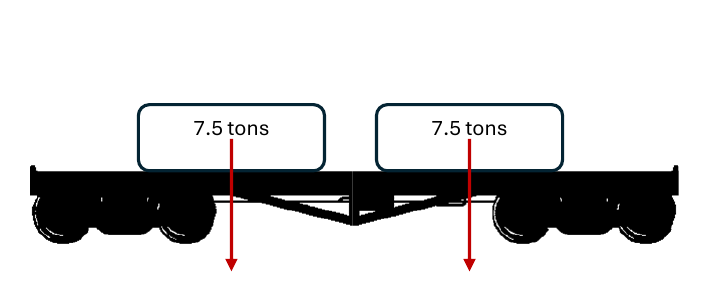}
        \caption{Half}
    \end{subfigure}\hfill
    \begin{subfigure}{0.33\textwidth}
        \includegraphics[width=\linewidth]{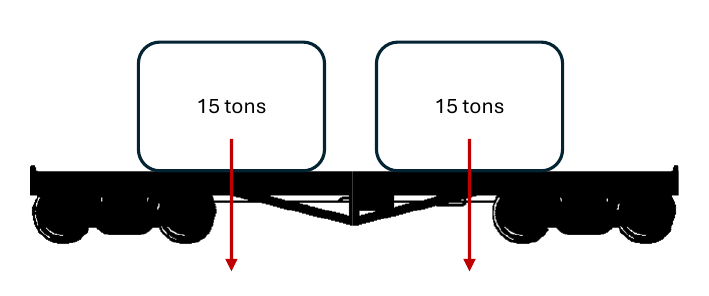}
        \caption{Full}
        \label{fig:fullscheme}
    \end{subfigure}\hfill
    \begin{subfigure}{0.33\textwidth}
        \includegraphics[width=\linewidth]{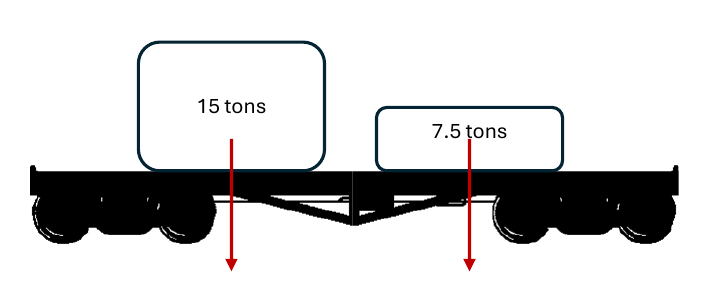}
        \caption{Unbalance 1}
    \end{subfigure}\hfill
    \begin{subfigure}{0.33\textwidth}
        \includegraphics[width=\linewidth]{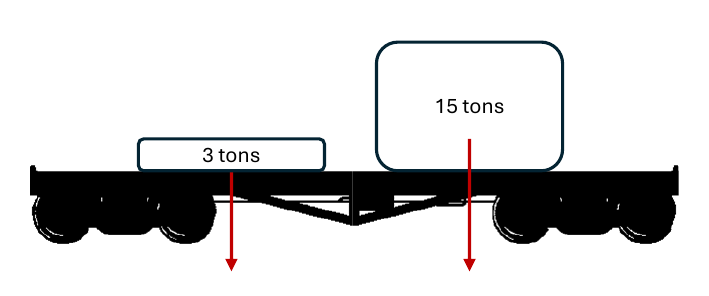}
        \caption{Unbalance 2}
    \end{subfigure}\hfill
    \begin{subfigure}{0.33\textwidth}
        \includegraphics[width=\linewidth]{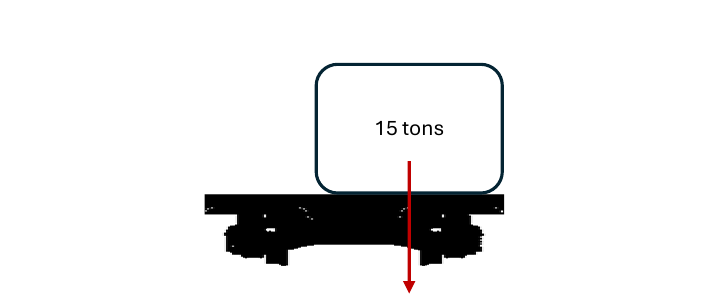}
        \caption{Unbalance 3}
    \end{subfigure}

    \caption{Load schemes}
    \label{fig:trainschemes}
\end{figure*}

\textbf{Polygonization.} For wheel polygonization, the periodic irregularity of the radial tread around the circumference of the wheel was considered by varying the wavelengths (\(\lambda\)) as a function of harmonic order (\(\theta\)) and wheel radius.
\begin{equation}
    \lambda = \frac{2\pi R_w}{\theta}, \theta = 1, 2, 3 \cdots, n
\end{equation}

\begin{figure}
    \centering

    \includegraphics[width=0.45\textwidth]{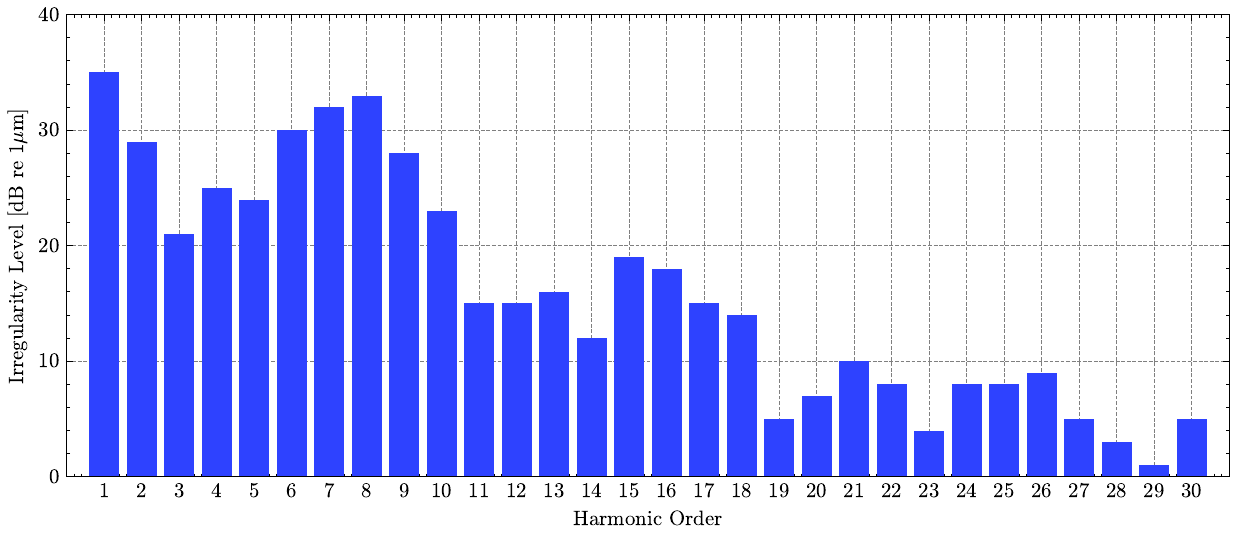}
    \caption{Amplitude (\(L_w\)) and harmonic (\(\theta\))}
    \label{fig:harmonic_order}
\end{figure}

The selected wheel profiles were characterized by the wavelengths in the first 20 harmonics, with the sixth to eighth harmonic orders being dominant, and different irregularity wheel profiles generated based on the sum of sine functions (\(H=20\)) as follows:
\begin{equation}
    w(x_w) = \sum_{\theta=1}^H A_\theta \sin \left( \frac{2\pi}{\lambda} x_w + \varphi_\theta \right),
\end{equation}
where \(A_\theta\) is the amplitude of the sine function for each wavelength, which is calculated by the function:
\begin{equation}
    A_{\theta} = \sqrt{2} \cdot 10^{\frac{L_{w}}{20}} \cdot w_{\text{ref}},
\end{equation}
with \(w_{ref} = 1\mu m\). The levels of wheel irregularity (\(L_w\)) were selected based on the irregularity spectrum in Figure~\ref{fig:harmonic_order}, produced with the measurement values of four wheels with polygonal damage \cite{https://doi.org/10.1155/2019/1538273}. Taking into account the phase angles to the sine functions that are uniformly and randomly distributed between \(0\) and \(2\pi\), several irregularity profiles of the wheel were generated to obtain different damage severities between 0.8 and 1.2 mm.

\textbf{Train profile.} Accurate identification of train types and their axle configurations is one of the most fundamental factors to effective axle counting and train detection systems (Figure \ref{fig:traintypes}). The way axles are arranged on a train depends largely on its design: whether it uses bogies, shared axles, or independent axle systems. The goal is to explore the key distinctions between different types of trains, focusing on how their axle layouts impact the process of counting axles and identifying vehicles in a railway network. The primary difference between bogie-equipped vehicles and axle-only vehicles is how the wheels are supported and arranged. Bogie vehicles use swiveling assemblies called bogies that hold multiple wheels and allow the train to handle curves and uneven tracks with greater ease. In contrast, axle-only vehicles have fixed axles that are shared between adjacent carriages. Moreover, such a difference in design is directly related to the way the wheels are arranged between consecutive carriages. Articulated trains have shared bogies for flexibility, conventional trains have independent bogies for stability, and regular trains have shared axles for simplicity. Each design serves specific purposes and is chosen based on the requirements of the rail system and the intended use of the train. However, such differing configurations of bogies and axles across articulated, conventional, and regular trains pose extreme environmental and operational variations that complicate pattern analysis for fault detection. To pertain such difficulties, the simulated data is comprised on a set of two types of trains: the Laagrss vehicle and the Alfa vehicle.

\textbf{Train load.} Figure~\ref{fig:trainschemes} illustrates the various train load schemes considered. The configurations include: (a) an empty train, where there is no load; (b) a half-loaded train with an equal load distribution of 7.5 tons; and (c) a fully loaded train with equal load distribution, applying 15 tons. Furthermore, three unbalanced load configurations are depicted: (d) Unbalance 1, where one side carries 15 tons while the other side carries 7.5 tons, forcing more on one of the sides, increasing the stress on those wheels; (e) Unbalance 2, where one side carries 15 tons and the other side carries a smaller load of 3 tons; and (f) Unbalance 3, where the entire load of 15 tons is concentrated on one side.

\textbf{Train speed.}  The Laagrss vehicle is designed for hauling freight, typically carrying heavier loads at slower speed. In contrast, the Alfa vehicle is a train built for passenger transport, prioritizing speed and lighter weight. For Laagrss, the configurations include passages in the range 40 to 120 km/h, and for Alfa in the range 40 to 220 km/h.

\textbf{Track profile.} In real circumstances, the rails will show small imperfections which affect the contact force values. Consequently, different irregularity profiles were generated for wavelengths between 1m and 30m, covering the D1 wavelength interval defined by the EN 13848-2 standard \cite{BSEN2003TrackGeometry}, with a sampling discretization of 1mm. The amplitude of the unevenness profile ranged from -2 mm to 2 mm, with a simulation length of 100 m. It is important to note that these wavelengths are significantly longer than those associated with wheel flat and polygonal damage. Consequently, the frequencies generated by track unevenness are much lower than those produced by defective wheels. More detailed descriptions of the unevenness profile generation process can be found in \cite{doi:10.1177/0954409719838115}.
 
\begin{figure}[!ht]
    \centering
    \begin{subfigure}[b]{0.95\columnwidth} 
        \centering
        \includegraphics[width=\linewidth]{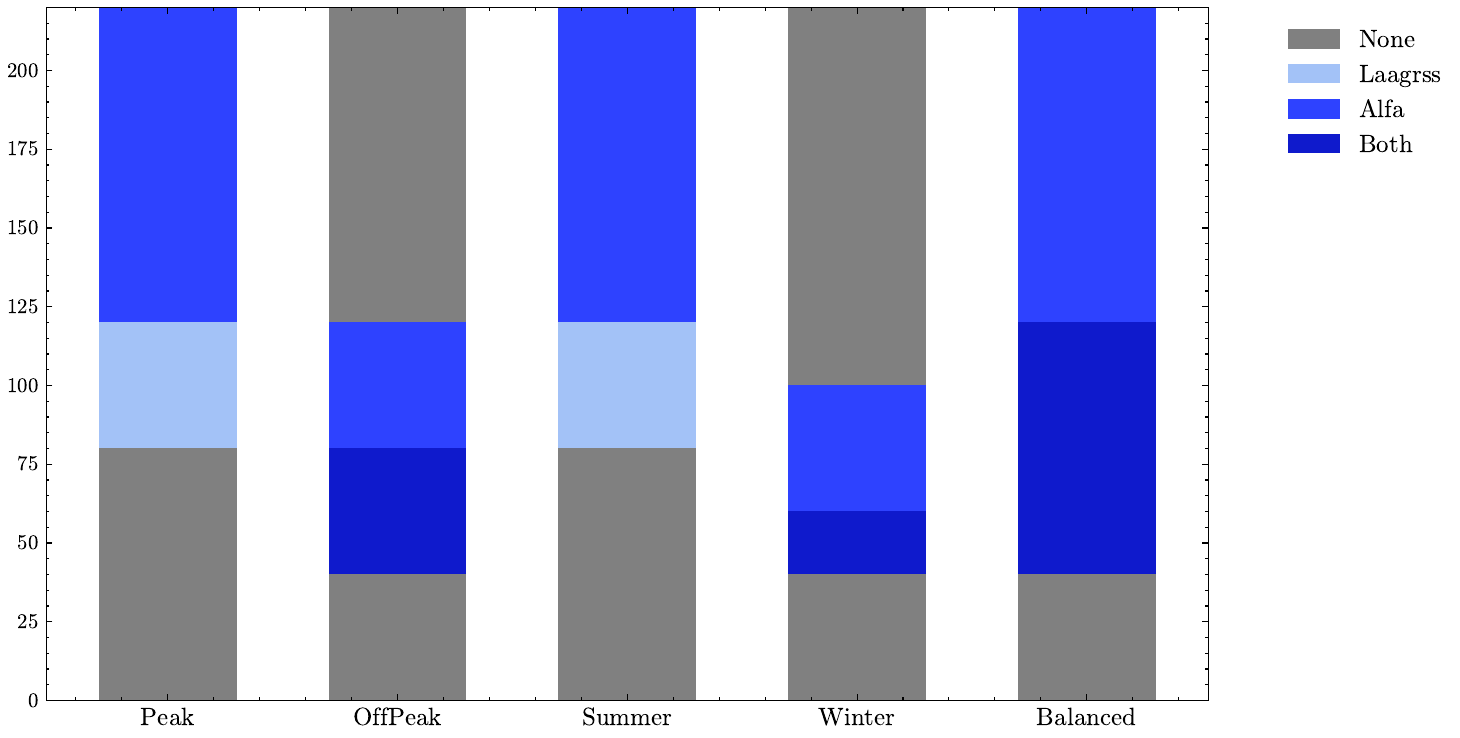}
        \caption{Speed (in km/h)}
        \label{fig:train_speed}
    \end{subfigure}
    \hfill
    \begin{subfigure}[b]{0.95\columnwidth} 
        \centering
        \includegraphics[width=\linewidth]{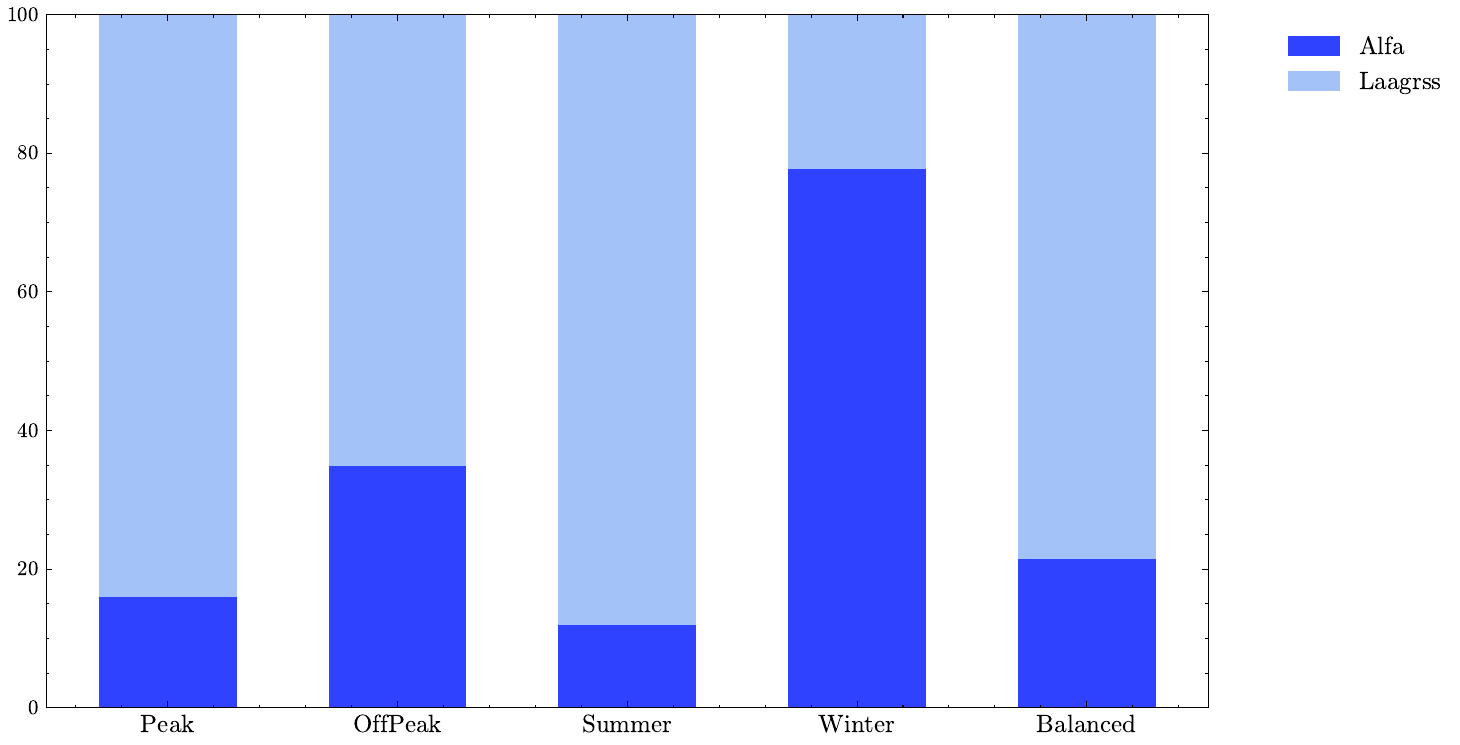}
        \caption{Train type}
        \label{fig:train_percentage}
    \end{subfigure}
    \begin{subfigure}[b]{0.95\columnwidth} 
        \centering
        \includegraphics[width=\linewidth]{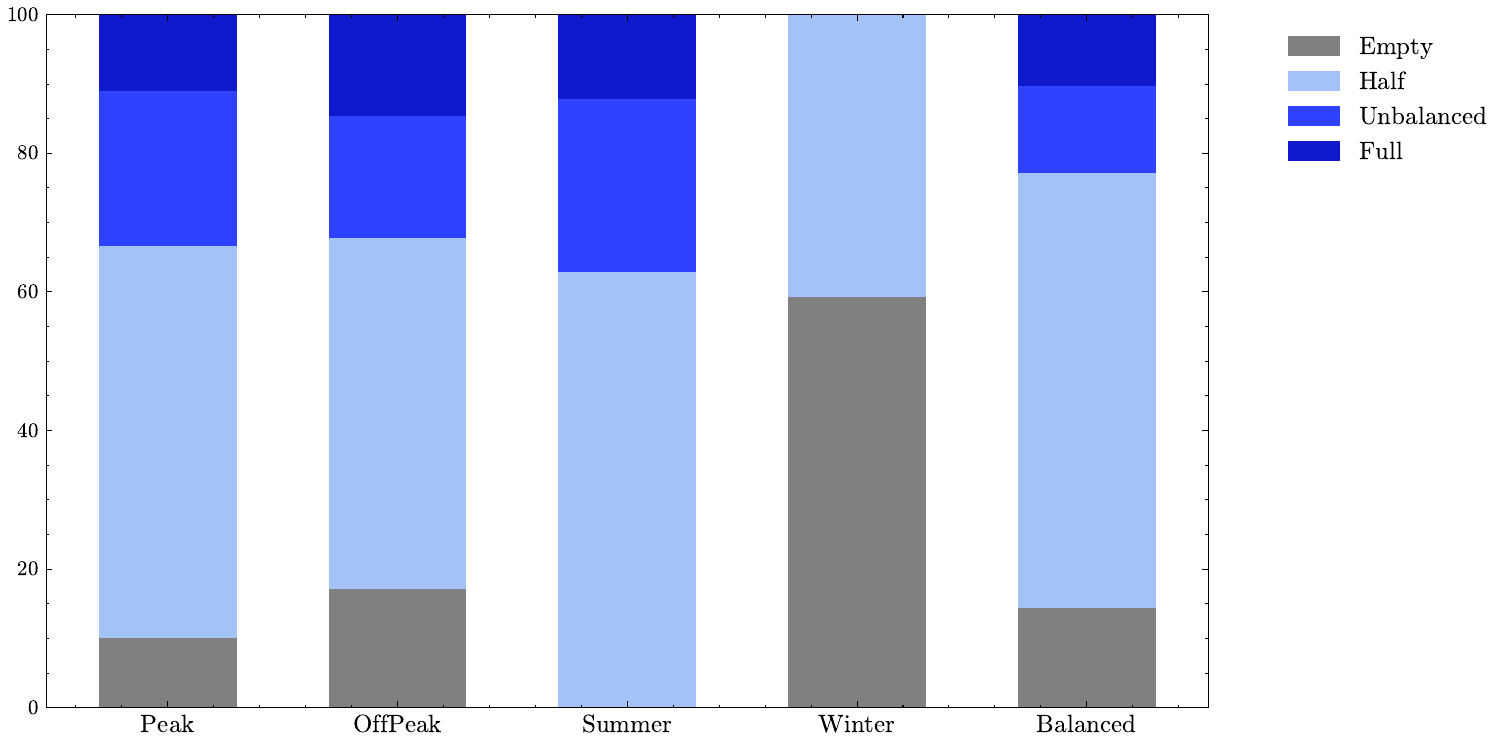}
        \caption{Cargo load}
        \label{fig:load_percentage}
    \end{subfigure}
    
    \caption{Distribution of EOVs across different train types}
    \label{fig:sample_train_partitioning}
\end{figure}

\textbf{Multi-domain data.} Moreover, to truly integrate aspects of continual learning in real-world applications, the training procedure must reflect potential temporal evolutions that honor the natural-time sequence. Consequently, various scenarios were created to represent possible temporal evolutions that respect the natural orderly sequence of time. These divisions are seasonal and recurrent, aimed at thoroughly testing the model's ability to retain past knowledge and manage new information, and illustrated in Figure~\ref{fig:sample_train_partitioning}, are:
(a) \textbf{Peak}: Comprise of higher speed trains, minimizing traffic disruptions during peak seasons; (b) \textbf{Off peak}: Features slower speed trains operating in a less congested railway setting, typical of off-peak seasons; (c) \textbf{Summer boom}: High flow of commerce and transport, characterized by high-speed trains with fully loaded wagons; (d) \textbf{Winter bust}: Featuring the slowest speeds and never completely filled wagons, indicative of a slow flow of transport and goods; (e) \textbf{Balanced}: Operating at medium speeds, optimizing for fuel efficiency and travel time, mantaining a steady flow of traffic.

\begin{figure*}[ht]
\centering
\begin{subfigure}{0.23\textwidth}
  \includegraphics[width=\linewidth]{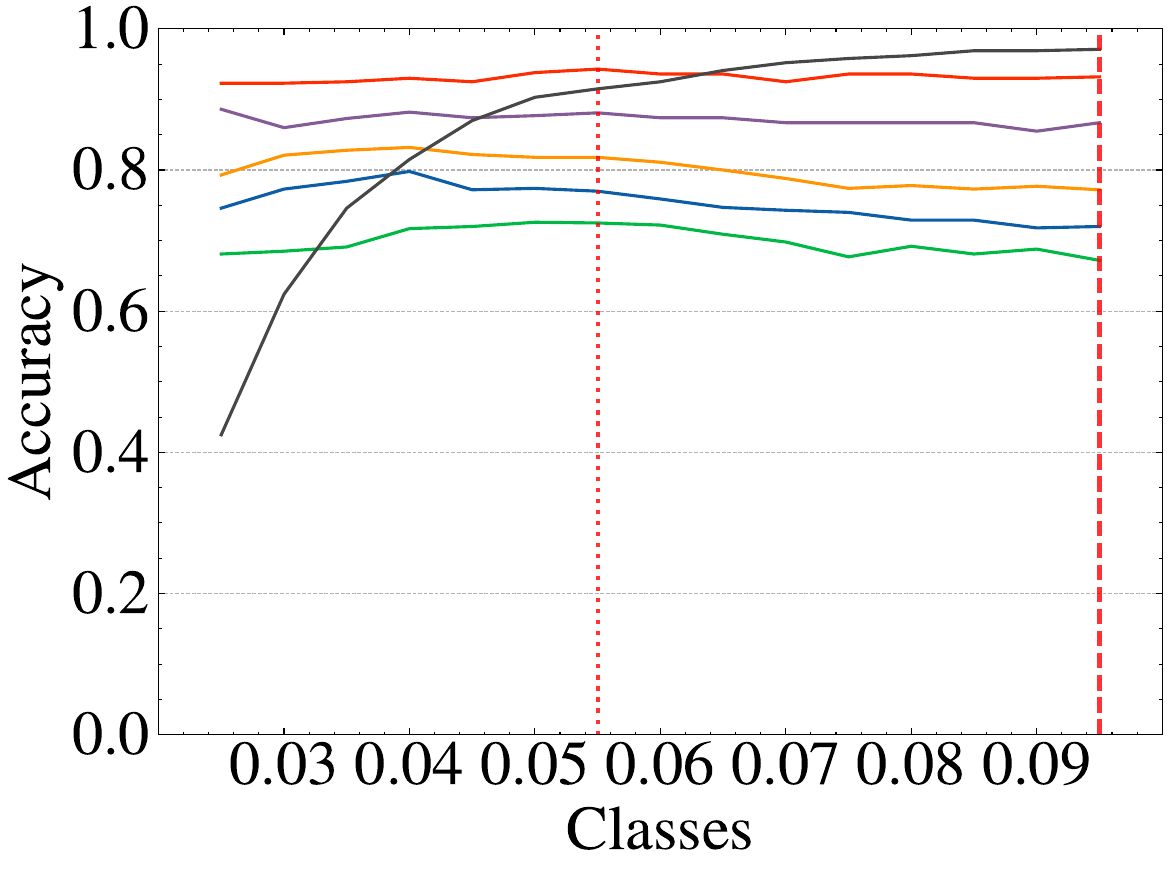}
  \caption{TonyBeltramelli (TB)}
\end{subfigure}\hfill
\begin{subfigure}{0.223\textwidth}
  \includegraphics[width=\linewidth]{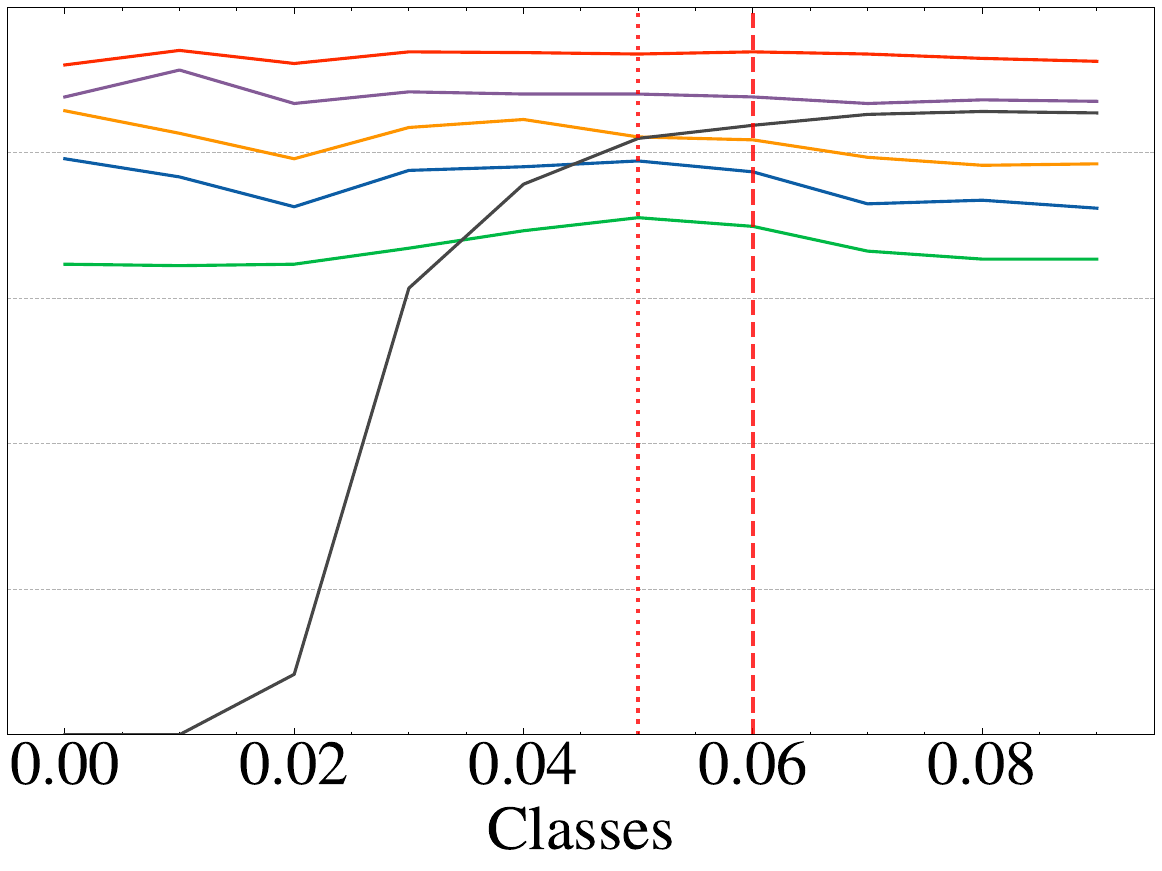}
  \caption{PeakDetect (PD)}
\end{subfigure}\hfill
\begin{subfigure}{0.223\textwidth}
  \includegraphics[width=\linewidth]{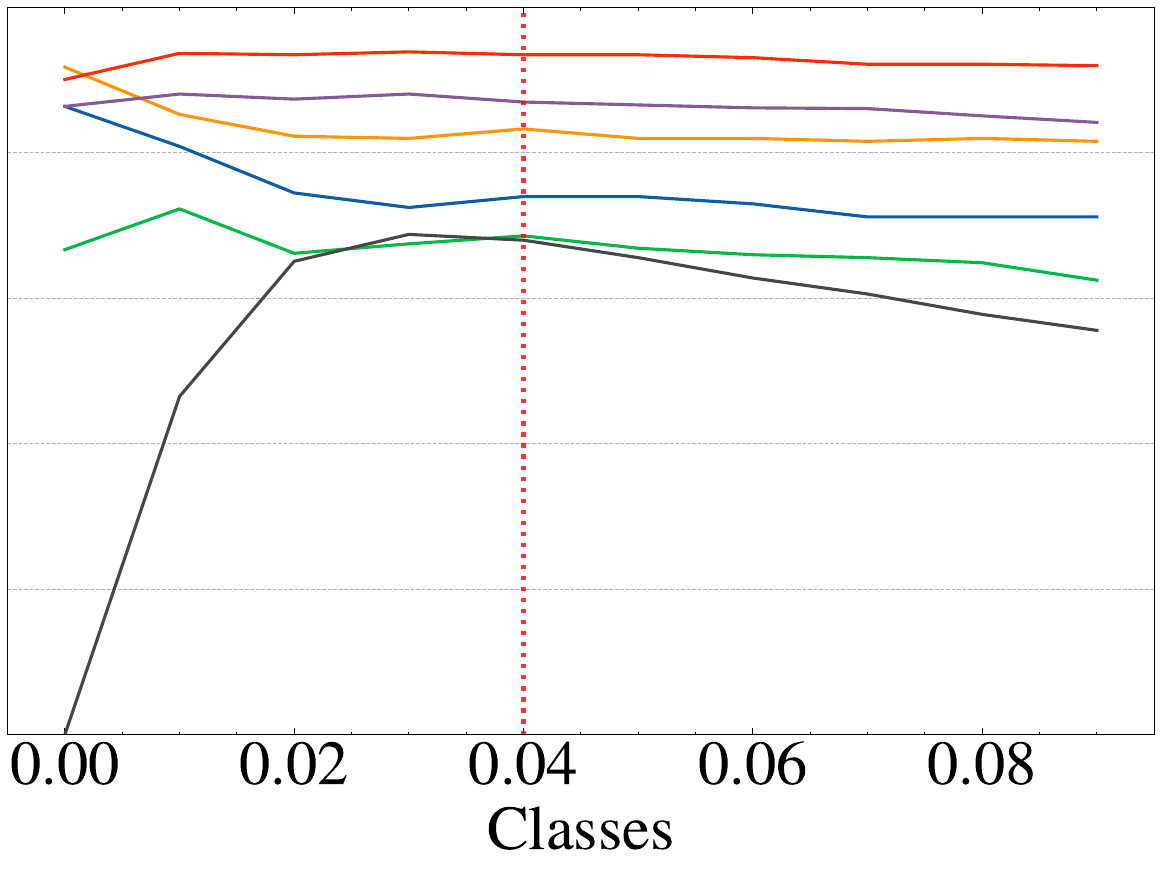}
  \caption{DetectPeak (DP)}
\end{subfigure}\hfill
\begin{subfigure}{0.314\textwidth}
  \includegraphics[width=\linewidth]{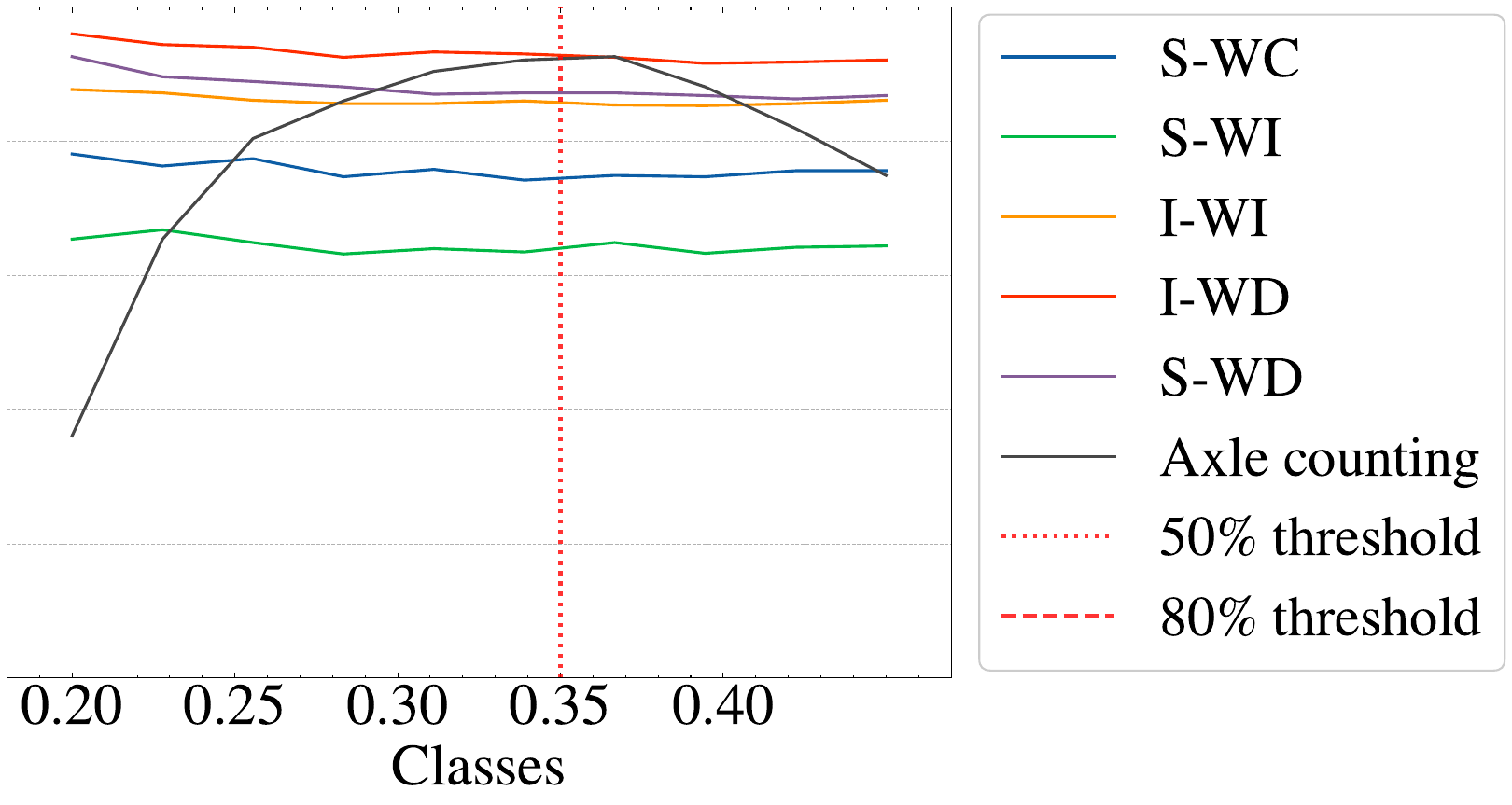}
  \caption{ScipyDetect (SD)}
\end{subfigure}
\caption{Axle counting strategies.}
\label{fig:ADACgraph}
\end{figure*}

\section{Experiments}
\label{sec:experiments}

To evaluate the effectiveness of the proposed methodology, two sets of quantitative experiments were performed.

On one hand, investigating the representational  quality of the VAE, and comparing how the semantic extraction and pre-processing strategies affect performance. For this purpose, a suite of batch evaluation metrics were considered: accuracy, i.e. ratio of correctly predicted instances to the total instances, precision, i.e. ratio of true positive predictions, recall, i.e. ratio of true relevant predictions, F1 Score, i.e. the harmonic mean of precision and recall, and AUC-ROC, i.e. the model's discrimination capability at various threshold settings. On the other hand, evaluating the semantics-enriched model's stability-plasticity trade-off in a continual learning setting, in terms of: forward transfer, intransigence, knowledge gain ratio, backward transfer, and forgetting. 

\begin{table}
\centering
\caption{Classification performance.}
\label{tab:performancevae_transposed}
\begin{tabular}{@{}lcccc@{}}
\toprule
\textbf{Model} & \textbf{Accuracy} & \textbf{Recall} & \textbf{Precision} & \textbf{F1 Score} \\
\midrule
Manual Features & 72\% & 74\% & 72\% &  75\% \\
\midrule
VAE & 95\% & 94\% & 91\% & 95\% \\
+ 1 Layer & + 4\% & + 5\% & + 8\% & + 3\% \\ 
+ 2 Layers & + 0\% & + 0\% & + 0\% & + 1\% \\
\bottomrule
\end{tabular}
\end{table}

\textbf{Representational quality.} The VAE was employed to generate a condensed, representative embedding of 20 points. This compressed representation aims to retain the signal's most informative features, serving as input to the downstream anomaly classifier. The encoder architecture consists of a fully connected linear layer with a ReLU (Rectified Linear Unit) activation function and an output size of 256. This is followed by a final linear layer with an output dimension of 40 (twice the latent space size) encoding the mean and logarithm of the variance of the latent distribution. A hyperparameter optimization process was conducted to minimize reconstruction loss on a validation set comprising 20\% of the data. Five parameters were explored across 50 trials: learning rate, latent space dimensionality, number of epochs, batch size, and KL divergence weight. The optimal configuration identified was a learning rate of 0.0005, latent dimension of 20, 150 epochs, batch size of 64, and KL divergence weight of 1.412. This configuration achieved a minimum validation loss of 683.01, in contrast to an average loss of 2702.43, corresponding to a 74.73\% improvement. To further investigate the suitability of the VAE architecture, an additional experiment was conducted to evaluate how different model depths affect downstream classification performance. Three configurations were tested while keeping the optimized hyperparameters fixed: (i) VAE with two hidden layers (output dimensions: 256, 64, 40), (ii) VAE with one hidden layer (output dimensions: 256, 40), and (iii) VAE with no hidden layers (direct projection to 40 dimensions). To contextualize the benefits of learned embeddings, a baseline was introduced using handcrafted features derived from traditional statistical and spectral analyses. These included measures such as mean, standard deviation, range, skewness, kurtosis, spectral energy, dominant frequencies, and Fast Fourier Transform (FFT) components. An optimized XGBoost classifier was used for all anomaly detection tasks. Table~\ref{tab:performancevae_transposed} presents the classification performance using features generated from each VAE configuration and the statistical baseline. The results clearly demonstrate the effectiveness of learned latent embeddings over handcrafted features. Among the tested architectures, the one-layer VAE offered the best compromise between model complexity and predictive performance. Based on these findings, the single-layer VAE was selected for the final model pipeline. This architecture was then used to generate a new dataset consisting of 40-dimensional embeddings per signal. For downstream tasks, these embeddings were concatenated with semantic features to further enrich the input representation.

\begin{table}
\caption{Confidence intervals for anomaly detection results by algorithm}
\centering
\begin{tabular}{ccccc}
\toprule
 & \textbf{TB} & \textbf{PD} & \textbf{DP} & \textbf{SD} \\ \midrule
S-WC & .70 ± .01 & .71 ± .01 & .71 ± .01 & .76 ± .01 \\ 
S-WC* & .82 ± .01 & .82 ± .01 & .80 ± .01 & .83 ± .01 \\ 
S-WI & .61 ± .01 & .62 ± .01 & .59 ± .01 & .54 ± .01 \\ 
S-WI* & .72 ± .01 & .72 ± .01 & .70 ± .01 & .69 ± .01 \\ 
I-WI & .76 ± .02 & .78 ± .01 & .76 ± .02 & .84 ± .02 \\ 
I-WI* & .86 ± .01 & .86 ± .01 & .85 ± .02 & .89 ± .02 \\ 
I-WD & .92 ± .01 & .92 ± .01 & .93 ± .01 & .90 ± .02 \\ 
I-WD* & \textbf{.93 ± .01} & \textbf{.93 ± .01} & \textbf{.94 ± .01} & \textbf{.93 ± .01} \\
S-WD & .88 ± .01 & .89 ± .01 & .86 ± .01 & .87 ± .01 \\ 
S-WD* & .90 ± .01 & .91 ± .01 & .89 ± .01 & .91 ± .01 \\ \bottomrule
\end{tabular}
\label{tab:resultsconfidence}
\end{table}

\textbf{Signal shape processing.} Before employing the aforementioned anomaly detection strategies, it is important to assess the isolated performance of the different axle counting systems, as well as as their sensitivity to hyperparameter selection. The sensitivity parameter of the peak detection algorithm governs its ability to identify peaks within the data. An increase in sensitivity corresponds to a decrease in the detection threshold, resulting in the identification of a greater number of peaks. Conversely, reducing sensitivity raises the threshold, leading to fewer peaks being detected. However, to fully investigate the influence of varying the signal shape detection algorithm, not only the individual performance of axle counting was addressed, but also the five aforementioned anomaly detection strategies, detailed in Table \ref{tab:ad_summary}. For the former, each train is annotated with its type, specifying axle layout, and evaluation performed by comparing detected peaks in strain gauge signals to the expected number and configuration of axles for each train passage. In particular, evaluation was performed on two levels: total axle count accuracy and axle grouping accuracy. For example, a Laagrss train can be expected to produce a signal reflecting 10 wheels: two isolated wheels at the beginning and end, with paired wheels in between. For the latter, the consequent impact on the anomaly detection performance was considered. As expected, Figure \ref{fig:ADACgraph} reveal that the axle counting system performance was notably affected by variations in peak detection sensitivity. As the sensitivity decreased, leading to a higher detection threshold, the system's accuracy showed a marked improvement. However, this performance enhancement reached a plateau, and in some cases a further reduction in sensitivity resulted in a slight decline in performance. This observation suggests the presence of an inflection point beyond which reducing sensitivity yields diminishing returns in terms of accuracy. In contrast, anomaly detection models exhibited an opposite trend, albeit with a more subtle effect. A decrease in sensitivity led to a decline in anomaly detection accuracy, with the most pronounced decrease observed in certain algorithms, such as Detectpeaks. Other algorithms displayed only minor fluctuations, indicating a differential sensitivity to the parameter changes.

\begin{figure*}[ht]
\centering
\begin{subfigure}{0.524\textwidth}
  \includegraphics[width=\linewidth]{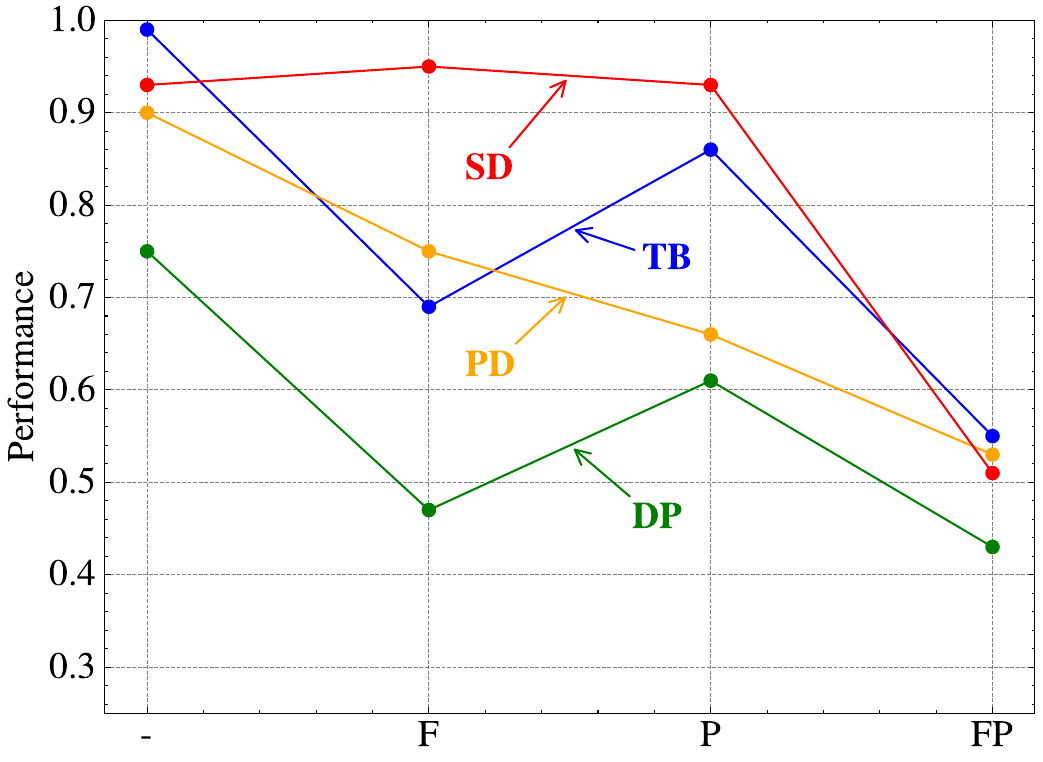}
  \caption{Anomaly Type}
\end{subfigure}\hfill
\begin{subfigure}{0.476\textwidth}
  \includegraphics[width=\linewidth]{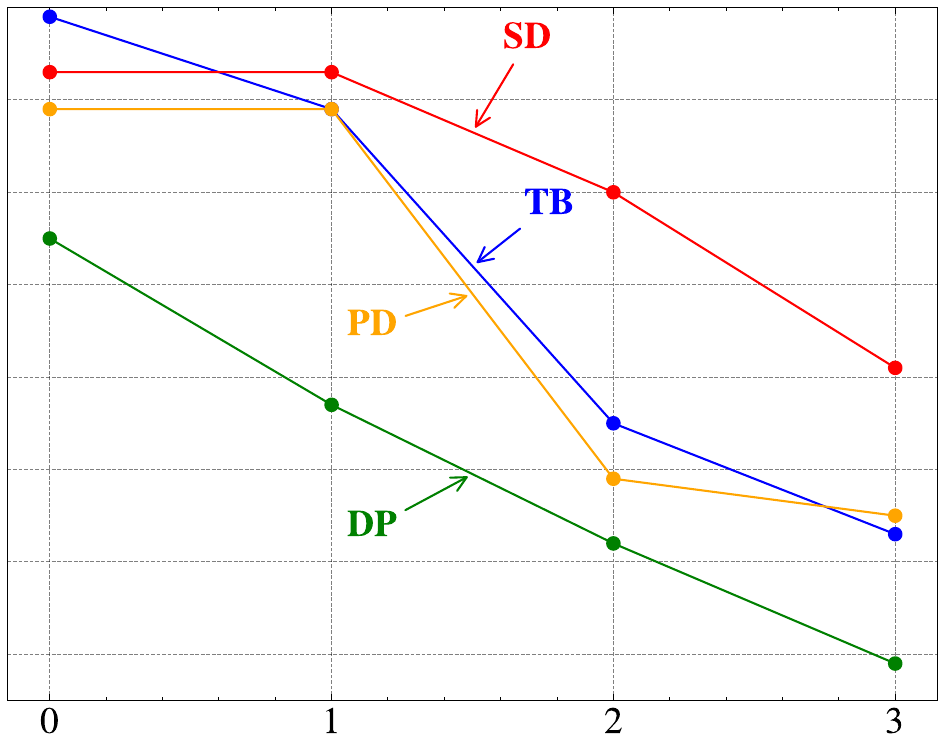}
  \caption{Anomaly Count}
\end{subfigure}\hfill
\caption{Comparison of peak detection algorithms under different fault conditions}
\label{fig:peakdetectionresults}
\end{figure*}

\textbf{Hyperparameter tuning.} Table \ref{tab:resultsconfidence} present the confidence intervals of the registered accuracies for the four tested algorithms Tony Beltramelli (\textit{TB}), Peakdetect (\textit{PD}), Detectpeaks (\textit{DP}), and Scipy (\textit{SD}) across the five aforementioned AD strategies, and their five counterparts which include additional information on train load and train speed. The anomaly detection classifier was tuned in each iteration to identify the optimal configuration, using a random search approach. The parameter ranges employed for tuning are: alpha (0.01-0.1), colsample\_bytree (0.5-1.0), subsample (0.5-1.0), learning\_rate (0.01-0.3), max\_depth (3-10), min\_child\_weight (1-6), and n\_estimators (50-200). The results from these iterations provided confidence intervals at a 95\% confidence level, allowing us to account for variability in model outcomes across different parameter settings. To determine the optimal sensitivity threshold for peak detection, two strategies were also employed. The first approach involved averaging the performance metrics of axle counting and anomaly detection, assigning equal weight to each task. The second approach put greater emphasis on anomaly detection by assigning it 80\% of the weight in the performance calculation. In most cases, the sensitivity thresholds identified through these two methods were consistent. However, when discrepancies appeared, the threshold that prioritized anomaly detection was selected. This approach ensured that the system's anomaly detection capability was maximized while maintaining an acceptable level of accuracy for axle counting. With these settings, Figure~\ref{fig:peakdetectionresults} summarizes the performance of the four peak detection algorithms across different anomaly types and anomaly multiplicities. The \textit{SD} method remains robust under both wheel flats and polygonization, whereas \textit{TB}, \textit{PD}, and \textit{DP} exhibit pronounced performance degradation for at least one of these fault types. Across all methods, a substantial loss in detection accuracy is observed when flats and polygonization co-occur within the same wheel passage. A closer analysis of the anomaly count suggests that this degradation is driven less by the specific fault morphology and more by the superposition of multiple anomalies, which distorts the signal shape and compromises the reliability of peak extraction.

\begin{table}[H]
\caption{Friedman test results (non-* / * values shown)}
\centering
\begin{tabular}{lcc}
\toprule
 & \textbf{Test statistic} & \textbf{p-value} \\
\midrule
TB & 59.813 / 58.453 & 3.176e-12 / 6.129e-12 \\
PD & 59.253 / 57.333 & 4.163e-12 / 1.053e-11 \\
DP & 59.253 / 58.453 & 4.163e-12 / 6.129e-12 \\
SD & 53.920 / 51.706 & 5.470e-11 / 1.589e-10 \\
\bottomrule
\end{tabular}
\label{tab:friedman_compact}
\end{table}

\begin{table}[H]
\caption{Shaffer pairwise test results (bold = p < 0.05; non-* / * values shown)}
\centering
\begin{tabular}{cccccccc}
\toprule
 &  & \textbf{S-WI} & \textbf{I-WI} & \textbf{I-WD} & \textbf{S-WD} \\ 
\midrule
\multirow{4}{*}{TB} 
 & S-WC & 1/1 & 1/1 & \textbf{,010/,010} & \textbf{,036/,029} \\ 
 & S-WI &  & \textbf{,036/,022} & \textbf{,010/,010} & \textbf{,010/,010} \\ 
 & I-WI &  &  & \textbf{,027/,054} & 1/1 \\ 
 & I-WD &  &  &  & 1/1 \\ 
\midrule
\multirow{4}{*}{PD} 
 & S-WC & 1/1 & 1/1 & \textbf{,010/,010} & \textbf{,022/,014} \\ 
 & S-WI &  & \textbf{,029/,044} & \textbf{,010/,010} & \textbf{,010/,010} \\ 
 & I-WI &  &  & \textbf{,036/,054} & 1/,727 \\ 
 & I-WD &  &  &  & 1/1 \\ 
\midrule
\multirow{4}{*}{DP} 
 & S-WC & 1/1 & 1/1 & \textbf{,010/,010} & \textbf{,044/,054} \\ 
 & S-WI &  & \textbf{,044/,034} & \textbf{,010/,010} & \textbf{,010/,010} \\ 
 & I-WI &  &  & \textbf{,022/,034} & 1/1 \\ 
 & I-WD &  &  &  & 1/1 \\ 
\midrule
\multirow{4}{*}{SD} 
 & S-WC & 1/1 & ,727/,819 & \textbf{,010/,010} & \textbf{,019/,036} \\ 
 & S-WI &  & \textbf{,010/,010} & \textbf{,010/,010} & \textbf{,010/,010} \\ 
 & I-WI &  &  & ,544/,805 & 1/1 \\ 
 & I-WD &  &  &  & 1/1 \\ 
\bottomrule
\end{tabular}
\label{tab:shaffercompact}
\end{table}

\begin{figure*}[h!]
\centering
\includegraphics[width=0.75\linewidth]{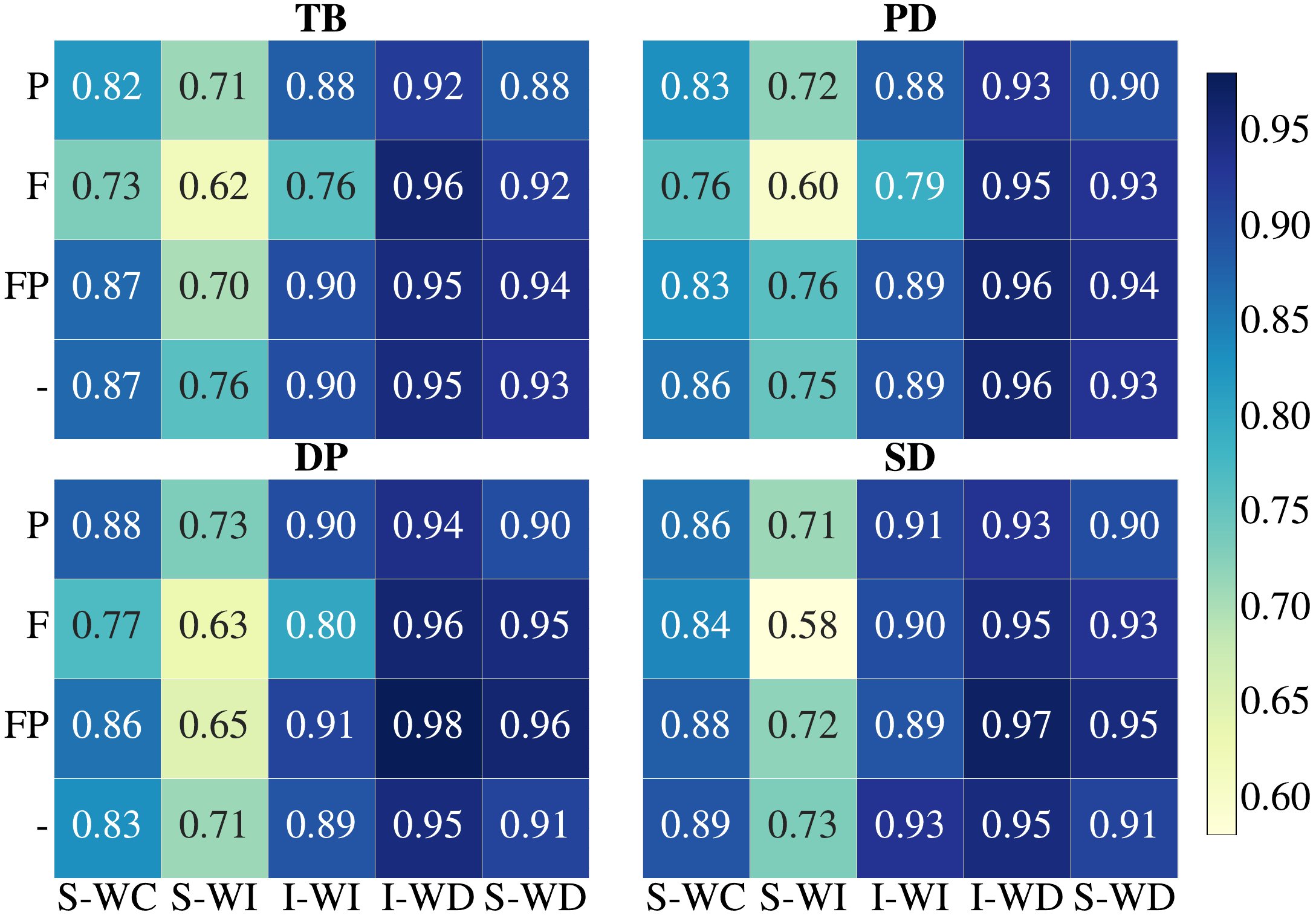}
\caption{Anomaly detection performance across the different semantic extraction strategies}
\label{fig:performanceTOPchucha}
\end{figure*}

\textbf{Statistical confirmation.} While such results may hint at a trend in performance, it is essential to substantiate this observation with formal statistical validation. For this purpose, both the Friedman and Shaffer tests were performed. The Friedman test is a non-parametric statistical test used to detect differences in treatments across multiple test attempts, which, unlike ANOVA, does not require the data to follow a specific distribution. The test evaluates if there are any statistically significant differences between the rankings of the treatments. If the null hypothesis is rejected, it implies that at least one of the scenarios is statistically different from the others, which allows for further post-hoc analysis to identify the best-performing scenarios. The results are presented in Table \ref{tab:friedman_compact}, which shows that there are statistically significant differences between the groups. The Friedman test statistics, which consistently range between 50 and 60, reflect a high degree of consistency in the ranking of scenarios across the different groups. This implies a robust distinction between the ranks of the experimental conditions, further supporting the conclusion that significant performance variations exist among the scenarios.However, the presence of overall significance does not imply that all pairs of scenarios exhibit statistically significant differences. It is possible that some groups may show comparable performance. To further investigate and pinpoint which specific scenarios differ from one another, a pairwise comparison using the Shaffer test was conducted, allowing one to simultaneously address the logical dependencies between the comparisons. Unlike the Nemenyi test, which assumes that all pairwise comparisons are independent, the Shaffer test adjusts for the number of comparisons to avoid overly conservative results. This makes the Shaffer test more powerful when comparing multiple groups, often leading to more significant findings when there are differences between scenarios. Its results are presented in Table \ref{tab:shaffercompact}. Notable significant differences were found primarily between S-WC and I-WD/S-WD, as well as between S-WI and I-WD/S-WD, across all algorithms. These findings suggest that the inclusion of semantic information in experiments I-WD and S-WD has a substantial effect, compared to the earlier conditions that lack such data. The general trend is consistent: incorporating semantic information (as seen in I-WD and S-WD) leads to improved model performance, with significant differences from earlier configurations that rely more heavily on signal data alone. Moreover, the Friedman test was conducted to assess whether incorporating train load and speed data led to statistically significant performance improvements (also displayed in Table \ref{tab:friedman_compact}). The results for all four algorithms yielded a Friedman test statistic of 0.000 and a p-value of 3.392e-06, which is well below the 0.05 significance threshold. Thus, confirming that additional semantic information consistently improves anomaly detection.

\textbf{Anomaly detection.} A correlation analysis was conducted to assess how the type and number of anomalies in each train passage influenced system performance. The correlation values were derived using the optimal threshold established in the hyperparameter tuning, ensuring that the analysis reflected the best conditions for both tasks. The analysis of the results reveals key patterns related to the impact of anomaly types: polygonization (P), flat wheels (F), multi-damage passages (F, P), and normal passages (-). The system performs best when either both types of anomalies are present or no anomalies are detected, suggesting that the system is more efficient in extreme cases, whether the signal is clear or severely disturbed. However, instances with a single anomaly, particularly flat wheels, show reduced performance, indicating that flat wheel anomalies are more difficult to detect. It is important to note that in Figure \ref{fig:peakdetectionresults} we concluded that as signal complexity increases, particularly in multi-damage passages with high anomaly counts, the axle counting system’s accuracy declines. The additional anomalies introduce more noise and disruptions, impairing the system's ability to detect and correctly identify axles. In contrast, anomaly detection benefits from more pronounced signal disturbances. While multi-damage trains are easier to detect for the anomaly detection system, these present significant challenges for axle counting, where precise structural recognition is crucial. These results underscore the complexity of balancing both tasks, especially in noisy and anomalous environments.

\begin{figure*}[ht]
\centering
\begin{subfigure}{0.50\textwidth}
  \includegraphics[width=\linewidth]{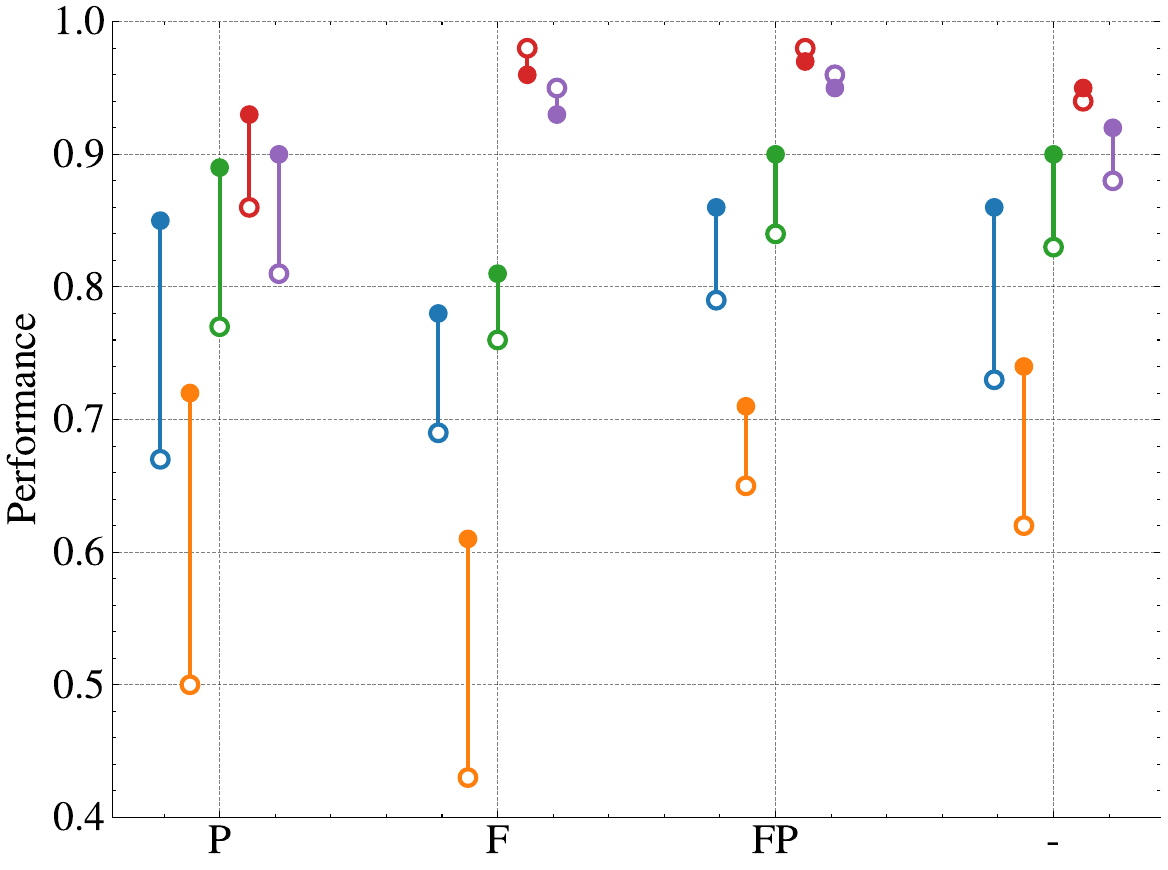}
  \caption{Anomaly Type}
\end{subfigure}\hfill
\begin{subfigure}{0.494\textwidth}
  \includegraphics[width=\linewidth]{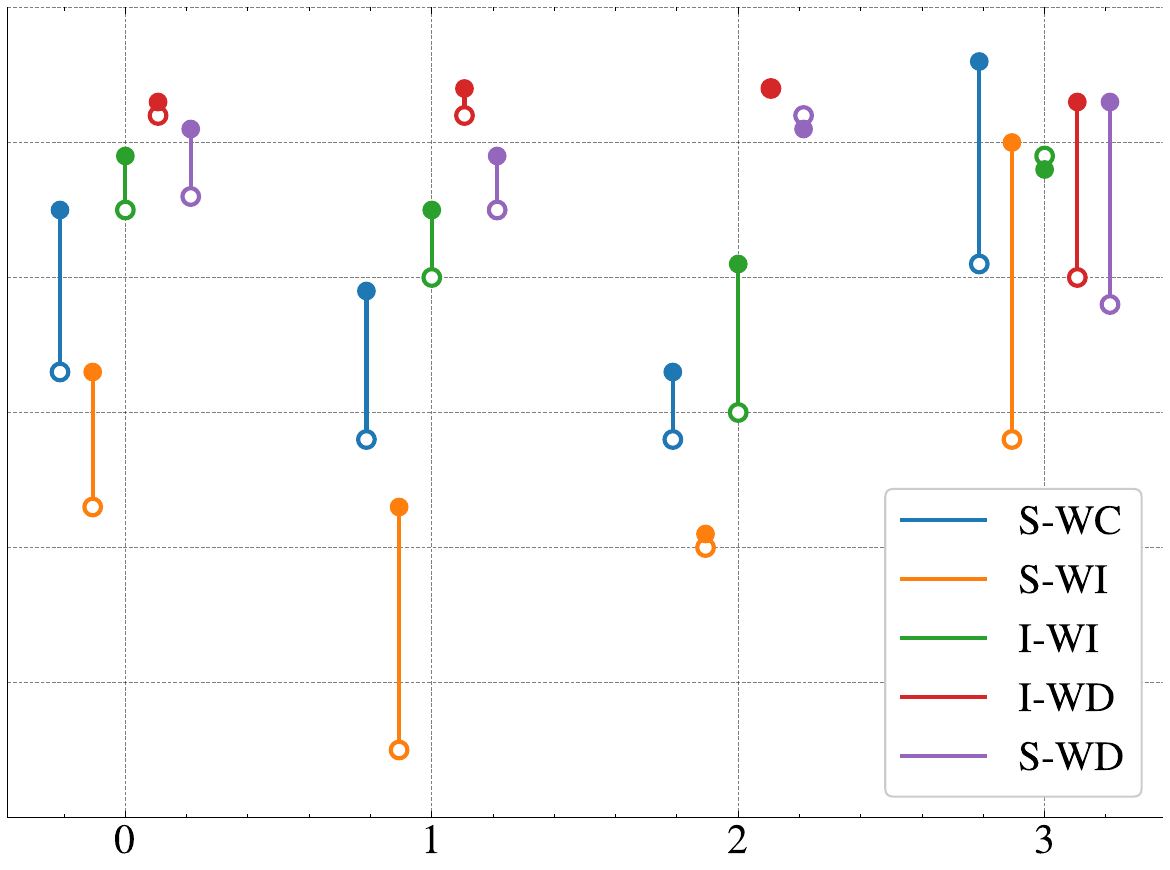}
  \caption{Anomaly Count}
\end{subfigure}\hfill
\caption{Impact of semantic enrichment on anomaly detection, averaged across the four peak detection algorithms. Colored markers correspond to semantic-augmented configurations (S-WC*, S-WI*, I-WI*, I-WD*, S-WD*), with contextual information on train load and speed. White markers denote baseline configurations (S-WC, S-WI, I-WI, I-WD, S-WD) described in Table \ref{tab:ad_summary}.}
\label{fig:performanceTOP}
\end{figure*}

\textbf{Semantics interacting with AD.} Figure \ref{fig:performanceTOP} examines the effect of semantics on anomaly detection performance across different anomaly types and quantities. As with anomaly type, the system achieves higher accuracy with either normal or multi-anomaly passages. Multi-damage passages generate more noise, enhancing anomaly detection. In contrast, single-anomaly passages produce subtler disturbances, reducing the system’s ability to differentiate and detect anomalies effectively. Across most strategies, we observe a substantial positive effect of including contextual information on train load and speed, with few exceptions, such as I-WD* and S-WD* in both (\textit{F}) and (\textit{F, P}).

\textbf{Continual learning.} To counteract the stability-plasticity trade-off, four different experience replay strategies were considered and compared against the naive retraining from scratch using the entire dataset every time new data arrives. RS consists of a memory sampler with reservoir sampling, where samples are selected and stored based on their true labels. LB consists of a loss-based memory sampler, where samples are selected and stored based on its associated loss value when compared to the true labels. P-RS and P-LB also respectively rely on reservoir sampling and loss-sampling, but samples are selected and stored based on the predicted labels from the past model, instead of the true labels. Each one one of these strategies was assessed for memory sizes of 200 or 800. Throughout these scenarios and different iterations, four different metrics were computed for assessing the continual learning performance (Table \ref{tab:ERmetricsNormalBold}):

\begin{itemize}
    \item \textit{Forward transfer} measures the influence of learning new tasks on the performance of previous tasks: 
    
    $\text{FWT} = \frac{1}{N-1} \sum_{i=1}^{N-1} \left( R_{0,i} - R_{i,i} \right)$,
    
    where \( R_{0,i} \) is the initial performance on task \( i \), and \( R_{i,i} \) is the final performance on task \( i \). Higher FWT indicates new tasks positively impacted previous ones.
    \item {Backward Transfer} measures the influence of learning new tasks on the performance of old tasks: 
    
    $\text{BWT} = \frac{1}{N(N-1)} \sum_{i=1}^{N-1} \sum_{j=i+1}^{N} \left( R_{j,i} - R_{i,i} \right)$,
    
    where \( R_{j,i} \) is the performance on task \( i \) after learning task \( j \), and \( R_{i,i} \) is the initial performance on task \( i \). Positive BWT values indicate learning new tasks has improved the performance on previous tasks. 
    \item \textit{Intransigence Measure} evaluates the difficulty a model has in learning new tasks: 
    
    $\text{IM} = \frac{1}{N} \sum_{i=1}^{N} \left( R_{\text{joint},i} - R_{\text{continual},i} \right)$,
    
    where \( R_{\text{joint},i} \) is the performance of the model trained jointly on all tasks, and \( R_{\text{continual},i} \) is the performance of the model trained continually. Lower IM values indicate better performance in learning new tasks.
    \item  \textit{Knowledge Gain Ratio} measures the proportion of knowledge retained from initial training:
    
    $\text{KGR} = \frac{1}{N} \sum_{i=1}^{N} \left( \frac{R_{\text{continual},i}}{R_{\text{initial},i}} \right)$,
    
    where \( R_{\text{initial},i} \) is the initial performance on task \( i \), and \( R_{\text{continual},i} \) is the final performance after continual learning on task \( i \). Higher KGR values indicate better retention and integration of knowledge.
\end{itemize}
 
\begin{table*}[ht]
\centering
\caption{Experience Replay impact on model performance for memory sizes 200 and 800}
\label{tab:ERmetricsNormalBold}
\begin{tabular}{@{}lcccccccccc@{}}
\toprule
\textbf{Metric} & \multicolumn{2}{c}{\textbf{Baseline}} & \multicolumn{2}{c}{\textbf{RS}} & \multicolumn{2}{c}{\textbf{P-RS}} & \multicolumn{2}{c}{\textbf{LB}} & \multicolumn{2}{c}{\textbf{P-LB}} \\
 & \textbf{200} & \textbf{800} & \textbf{200} & \textbf{800} & \textbf{200} & \textbf{800} & \textbf{200} & \textbf{800} & \textbf{200} & \textbf{800} \\
\midrule
\textbf{Forward Transfer} & -0.0109 & -0.0102 & 0.0934 & 0.0377 & 0.1882 & 0.0947 & 0.1492 & 0.2427 & \textbf{0.3122} & 0.1414 \\
\textbf{Backward Transfer} & -0.0117 & -0.0038 & -0.0129 & -0.0219 & -0.1170 & -0.1136 & -0.0077 & \textbf{-0.0018} & -0.1164 & -0.1240 \\
\textbf{Intransigence Measure} & \textbf{0.0104} & 0.0111 & 0.1086 & 0.0628 & 0.1990 & 0.1031 & 0.2192 & 0.3038 & 0.3637 & 0.1864 \\
\textbf{Knowledge Gain Ratio} & 3.1971 & 3.2005 & 2.8668 & \textbf{3.0277} & 2.5662 & 2.8927 & 2.4983 & 2.2169 & 2.0126 & 2.6144 \\
\bottomrule
\end{tabular}
\end{table*}

The findings indicate that while RS methods performance deteriorates with reduced memory size due to its non-selective sampling approach, LB methods outperform under similar conditions. LB methods effectively prioritize more informative samples, allowing the model to maintain performance even in resource-constrained environments. In particular, increasing memory size generally improves the model's ability to transfer knowledge to new tasks while reducing the negative impact on previously learned tasks. Both P-RS and P-LB demonstrate strong forward learning, yet also reveal issues related to forgetting, suggesting the need for strategies that enhance retention. Among these, LB benefits the most from a larger memory buffer, indicating an optimal balance between learning and retention.

\section{Conclusion}
\label{sec:conclusion}

In this work, a label-efficient continual learning framework is proposed to jointly model the passage of train axles using (1) learned latent representations from accelerometer data, which are highly sensitive to wheel–rail damage, and (2) real-time Fiber Bragg Grating (FBG) strain measurements, whose stable temporal structure helps mitigate the effects of non-linear operational and environmental variability. A key practical advantage of the proposed predictive maintenance system is that both the wayside sensors and the connecting optical fibers are passive and inherently immune to electromagnetic interference (EMI), requiring no electrical power along the trackside up to the head-end interrogation unit, which may be located tens of kilometers from the measurement points. This property is particularly advantageous in modern electrified railway systems, where EMI severely impacts legacy sensing solutions. By combining data-driven fault sensitivity with EMI-resilient semantic axle characterization, the proposed approach offers a robust alternative to traditional axle counters, reducing the need for additional trackside equipment, maintenance operations, and dedicated data processing infrastructure.

\textbf{Extensive ablations.} The experimental results demonstrated the robustness of the Scipy (\textit{SD}) peak detection algorithm, which achieved consistently high axle counting accuracy across different train types, reaching up to 93\% in scenarios enriched with semantic information (S-WD* and I-WD*). This highlights that FBG sensors provide a non-intrusive and reliable means of detecting train axles and monitoring wheel conditions, even in EMI-prone electrified environments. Furthermore, the combination of accelerometer-derived latent embeddings with strain gauge-based semantic features substantially improved anomaly detection performance. Across all tested algorithms, semantic enrichment increased detection accuracy by 5–12 percentage points on average, with the highest gains observed for multi-anomaly passages. In continual learning scenarios, replay-based strategies, particularly loss-based memory sampling (LB) with larger memory buffers (800 samples), achieved forward transfer values of 0.24–0.31 and reduced forgetting compared to naive replay, confirming that the integration of both data modalities allows the system to maintain high predictive performance while adapting to new data.

\textbf{Future work.} While the results are promising, this preliminary study is not yet sufficient to guarantee the reliability of the proposed FBG-based axle-counting system in real-world railway environments. Further on-track testing and extensive measurements are necessary to mature this technology for practical adoption in the highly conservative railway industry. Future research should focus on: (i) developing more intelligent peak detection algorithms to improve accuracy and robustness, (ii) implementing redundant FBG sensors at each measurement point to increase availability and reliability, and (iii) investigating the sensitivity of FBG sensors to thermal expansion and environmental variations. Additionally, broader studies encompassing diverse operational and environmental conditions are needed to refine domain-specific strategies, optimize sensor placement, and enhance data acquisition. Integration with other railway sensing applications could further leverage the semantic-enriched continual learning framework, potentially providing richer operational insights at a relatively low cost compared with conventional axle-counting technologies. 

\section*{Acknowledgments}

Funded by Portuguese Foundation for Science and Technology under Ph.D. scholarship PRT/BD/154713/2023 and project doi.org/10.54499/UID/00760/2025.

\bibliographystyle{cas-model2-names}

\bibliography{cas-refs}


\end{document}